\documentclass{article} 
\usepackage[preprint]{colm2025_conference}
\usepackage{asymptote}
\usepackage{adjustbox}
\usepackage{microtype}
\usepackage{hyperref}
\usepackage{url}
\usepackage{booktabs}
\usepackage{graphicx} 
\usepackage{lineno}
\usepackage{wrapfig}
\usepackage{enumitem}
\usepackage[many,breakable]{tcolorbox}
\usepackage{lipsum}
\usepackage{pgfplots}
\usepackage{tikz}
\usepackage{tcolorbox}
\tcbuselibrary{skins}
\usepackage{float}
\usepackage{soul}
\usepackage{tablefootnote}

\usepackage[utf8]{inputenc}
\usepackage{amsmath}
\usepackage{amssymb}
\usepackage{colortbl}
\usepackage{array}
\usepackage{multirow}

\definecolor{codegreen}{rgb}{0,0.6,0}
\definecolor{codegray}{rgb}{0.5,0.5,0.5}
\definecolor{codepurple}{rgb}{0.58,0,0.82}
\definecolor{backcolour}{rgb}{0.95,0.95,0.92}

\definecolor{darkred}{rgb}{0.6,0.0,0.0}
\definecolor{darkgreen}{rgb}{0,0.50,0}
\definecolor{darkblue}{rgb}{0, 0, 0.5}
\definecolor{lightblue}{rgb}{0.0,0.42,0.91}
\definecolor{orange}{rgb}{0.99,0.48,0.13}
\definecolor{grass}{rgb}{0.18,0.80,0.18}
\definecolor{pink}{rgb}{0.97,0.15,0.45}

\hypersetup{colorlinks=true, citecolor=darkblue, linkcolor=darkblue, urlcolor=darkblue}

\usepackage{listings}
\lstdefinestyle{mystyle}{
    backgroundcolor=\color{backcolour},   
    commentstyle=\color{codegreen},
    numberstyle=\tiny\color{codegray},
    stringstyle=\color{codepurple},
    basicstyle=\ttfamily\footnotesize,
    breakatwhitespace=false,         
    breaklines=true,                 
    captionpos=b,                    
    keepspaces=true,                 
    numbers=left,                    
    numbersep=5pt,                  
    showspaces=false,                
    showstringspaces=false,
    showtabs=false,                  
    tabsize=2
}

\usepackage{amssymb}%
\usepackage{pifont}%
\newcommand{\cmark}{\textcolor{green}{\ding{51}}}
\newcommand{\xmark}{\textcolor{red}{\ding{55}}}

\definecolor{darkgreen}{RGB}{0,150,0}
\definecolor{darkred}{RGB}{150,0,0}

\definecolor{lightgray}{gray}{0.9}
\sethlcolor{lightgray}

\newtcolorbox{smallredboxes}{
  enhanced,
  colback=red!10, 
  boxrule=0pt,
  frame hidden,
  borderline={0.5mm}{0mm}{red!25}, 
  arc=5mm, 
  boxsep=3mm, 
  left=3mm, 
  right=3mm, 
  fontupper=\small, 
  width=0.9\textwidth 
}

\newtcolorbox{smallblueboxes}{
  enhanced,
  colback=blue!10,
  boxrule=0pt,
  frame hidden,
  borderline={0.5mm}{0mm}{blue!25},
  arc=5mm, 
  boxsep=3mm, 
  left=3mm,   
  right=3mm,  
  fontupper=\small, 
  width=0.9\textwidth 
}

\newtcolorbox{smallorangeboxes}{
  enhanced,
  colback=orange!10, 
  boxrule=0pt,
  frame hidden,
  borderline={0.5mm}{0mm}{orange!25}, 
  arc=5mm, 
  boxsep=3mm, 
  left=3mm, 
  right=3mm, 
  fontupper=\small, 
  width=0.9\textwidth 
}

\title{Climbing the Ladder of Reasoning: What LLMs Can—and Still Can't—Solve after SFT?}

\author{
Yiyou Sun$^1$\thanks{Corrsponding to \texttt{\{sunyiyou,dawnsong\}@berkeley.edu}}, Georgia Zhou$^1$, Haoyue Bai$^2$, Hao Wang$^1$, Dacheng Li$^1$, Nouha Dziri$^3$, Dawn Song$^1$\\
$^1$University of California, Berkeley,
$^2$University of Wisconsin, Madison,
$^3$Allen Institute for AI\\
\vspace{-0.4cm}
}

\begin{document}

\ifcolmsubmission
\linenumbers
\fi

\maketitle

\begin{abstract}
Recent supervised fine-tuning (SFT) approaches have significantly improved language models' performance on mathematical reasoning tasks, even when models are trained at a small scale. However, the specific capabilities enhanced through such fine-tuning remain poorly understood. In this paper, we conduct a detailed analysis of model performance on the AIME24 dataset to understand how reasoning capabilities evolve. We discover a ladder-like structure in problem difficulty, categorize questions into four tiers (Easy, Medium, Hard, and Extremely Hard (Exh)), and identify the specific requirements for advancing between tiers. We find that progression from Easy to Medium tier requires adopting an R1 reasoning style with minimal SFT (500-1K instances), while Hard-level questions suffer from frequent model's errors at each step of the reasoning chain, with accuracy plateauing at ~65\% 
 despite logarithmic scaling. Exh-level questions present a fundamentally different challenge; they require unconventional problem-solving skills that current models uniformly struggle with. Additional findings reveal that carefully curated small-scale datasets offer limited advantage—scaling dataset size proves far more effective. Our analysis provides a clearer roadmap for advancing language model capabilities in mathematical reasoning. The code repository is available at \url{https://github.com/sunblaze-ucb/reasoning_ladder.git}.
\end{abstract}
\vspace{-0.4cm}

\section{Introduction}
\label{sec:intro}
\vspace{-0.4cm}

\begin{figure}[htb]
    \centering
    \includegraphics[width=0.97\linewidth]{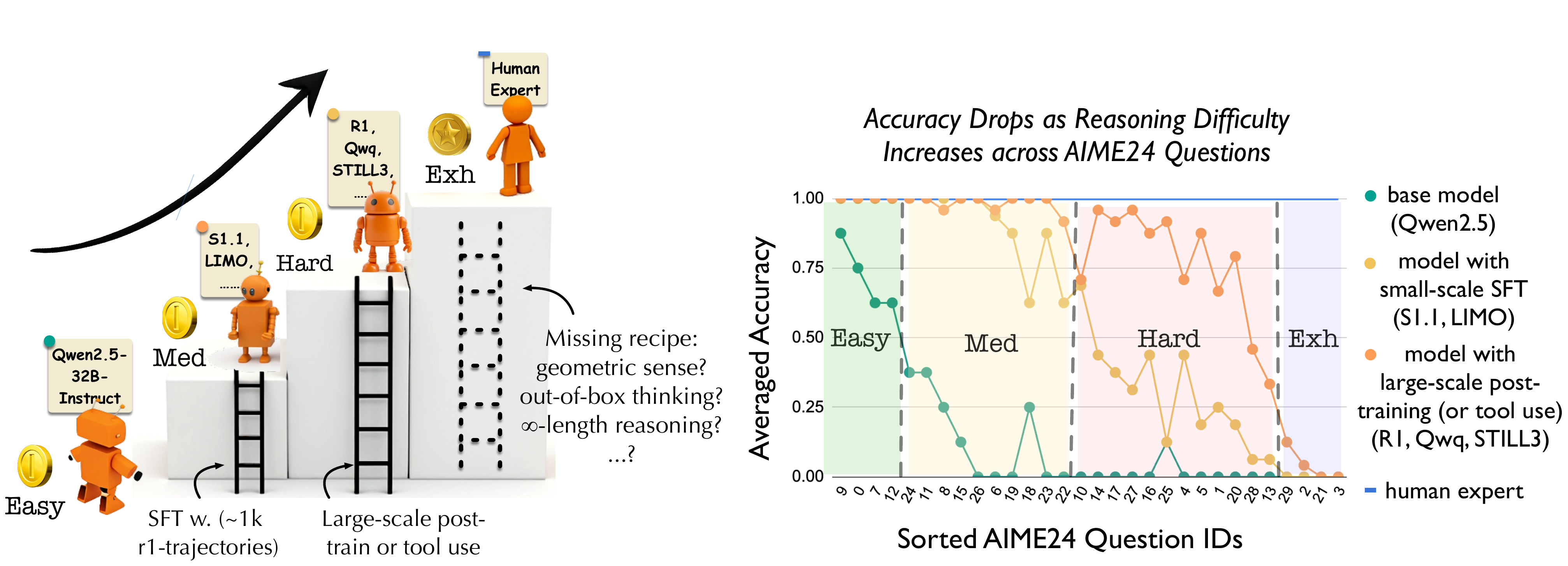}
    \caption{\footnotesize \textbf{Climbing the Reasoning Ladder on AIME24}.
\textbf{Left}: A conceptual illustration of what it takes for a base model to tackle increasingly difficult problems (Easy → Med → Hard → Exh) on the AIME24 benchmark. Model performance improves from the \textit{Qwen2.5-32B-Instruct}, to small-scale SFT (1k R1-trajectory SFT), and further to large-scale post-training or tool-augmentation. However, the strongest models still fall short of human expert performance at the most challenging level.
\textbf{Right}: Averaged accuracy across AIME24 question IDs, sorted by increasing overall difficulty (as determined by the average accuracy across six models: \textit{Qwen2.5-32B-Instruct}~\citep{qwen2.5}, \textit{S1.1-32B}~\citep{muennighoff2025s1}, \textit{LIMO-32B}~\citep{ye2025limo}, Deepseek-\textit{R1}~\citep{guo2025deepseek}, \textit{Qwq-32B}~\citep{qwq32b}, and \textit{STILL3-32B}~\citep{chen2025still}). Each model attempts each question 8 times with averaged accuracy. Colored lines represent the mean performance for each model category.
}
    \label{fig:teaser}
\end{figure}
For reasoning tasks, even relatively small-scale supervised fine-tuning (SFT) approaches, like LIMO~\citep{ye2025limo} and s1~\citep{muennighoff2025s1}, can markedly improve a model’s performance on mathematical problems (see Table~\ref{tab:math_improvement}). 
However, do these models risk overfitting the test set, or can they truly generalize?  If they do generalize, precisely what capabilities are enhanced through small-scale SFT, and which limitations persist? Although these models often excel on popular benchmarks, our understandings of their specific strengths and weaknesses remains incomplete.

Beyond aggregating scores on standard benchmarks, the broader impact of SFT is not yet well understood. For instance, which types of questions—previously unsolved by the base model—become solvable with reasoning-based SFT? Are these improvements merely tied to the exact trajectories seen during training, or do they reflect a deeper transfer of problem-solving strategies, such as the coordinate-based techniques used in geometry? Furthermore, what types of questions remain challenging or unsolved despite sufficient SFT? Prior research~\citep{muennighoff2025s1,ye2025limo} has offered early insights, suggesting that factors like correctness, solution length, and response diversity play a non-trivial role in SFT. Yet a more granular investigation is needed to fully determine how these models evolve through SFT, and to identify remaining gaps that must be addressed.

To investigate these questions, we analyze the AIME24 dataset \citep{aime2024problem}, selected for its complexity, diversity, and widespread adoption as a primary evaluation metric in recent reasoning-focused research \citep{openthoughts,muennighoff2025s1,ye2025limo}. Our examination reveals a ladder-like structure in AIME24 questions, wherein models that reliably solve higher-tier questions generally succeed on lower-tier ones (see Figure~\ref{fig:teaser}). Motivated by this observation, we categorize AIME24 questions into four tiers—Easy, Medium, Hard, and Exh—and analyze the specific requirements for advancing from one tier to the next.

\begin{itemize}[leftmargin=0pt, itemindent=10pt] 
\item Easy-level questions are typically solvable by base models without additional tuning. We find that progressing from Easy-level to Medium-level proficiency ($>$90\% average accuracy) primarily requires adopting an R1 reasoning style and long inference context. The minimal condition for SFT in this transition is approximately 500-1K instances of R1-style \footnote{ In paper, we use \texttt{R1-style} refers to problem-solving traces featuring extended chain-of-thought reasoning with self-reflection mechanisms, characterized by substantial length and explicit verification steps that allow models to check and correct their own work.} trajectory data for solving math questions, \textit{regardless of their specific categories}.

\item When advancing to Hard-level questions, an R1-like reasoning style alone proves insufficient. The main obstacle becomes intrinsic instability in deeper exploration and heavier computational demands. Performance improvement at this level follows a logarithmic scaling law over the size of the SFT dataset, with accuracy plateauing at $\sim$65\% on Hard-level questions. 

\item Exh-level questions pose a fundamentally different challenge, characterized by their dependence on unconventional strategies. These strategies often require out-of-the-box insights or strong geometric intuition. Current models uniformly struggle at this level, indicating fundamental limitations that we discuss thoroughly in Section~\ref{sec:hard-to-exp}.

\end{itemize}

Our analysis also yields additional important insights for future research:
\vspace{-0.2cm}
\begin{enumerate}[leftmargin=0pt, itemindent=10pt] 
\item \textbf{Potential vs. stability}. Models with small-scale SFT demonstrate the potential to solve as many AIME24 questions as Deepseek-R1 when given multiple attempts, but their overall accuracy remains significantly lower due to instability in deep exploration and computation.

\item \textbf{Careful curation of small-scale SFT datasets yields marginal gain}. Performance across various math categories remains consistent within a narrow range (55±4\%), with even specifically constructed similar dataset and randomly constructed dataset showing only marginal performance differences of about 1\%.

\item \textbf{Scaling SFT dataset remains important}. This finding contradicts recent claims that very small datasets ($\sim$1K samples) are sufficient and better~\citep{muennighoff2025s1,ye2025limo}. However, adding more examples yields diminishing benefits on Hard-level problems, indicating a performance plateau.

\item \textbf{Higher-level intelligence barriers}. Models trained using SFT tend to adopt similar solution strategies, raising fundamental questions about whether higher-level reasoning capabilities can be developed through SFT alone.

\end{enumerate}

Overall, by identifying the specific challenges at each difficulty level and the conditions under which they are mitigated or remain unresolved, we provide a clearer roadmap for advancing LLM-based mathematics and reasoning capabilities.

\section{Climbing the ladder of reasoning}
\label{sec:ladder}
Recent studies~\citep{muennighoff2025s1, ye2025limo} have shown that training a 32B base model using supervised fine-tuning (SFT) on a relatively small set of reasoning trajectories can yield impressive results. Notably, such models can outperform \texttt{o1-preview}~\citep{openai2024learning} on widely used math benchmarks like MATH500~\citep{hendrycks2measuring}, achieving 89\% accuracy compared to 81.4\% with \texttt{o1-preview} (Table~\ref{tab:math_improvement}). These models also demonstrate a degree of generalization on challenging datasets such as GSM-Plus~\citep{li2024gsm}, which includes perturbed problems, and HLE~\citep{phan2025humanity}, which features particularly difficult math tasks (see Appendix~\S\ref{sec:sup_generalization}).

These findings raise a natural and important question: What capabilities are gained through SFT with reasoning trajectories and what limitations remain to be addressed? 
In this section, we address these questions by systematically exploring the reasoning ladder using the AIME2024 dataset. Before delving into our analysis, we introduce the experimental setup used throughout our study.

\subsection{Experimental Setup} 
\textbf{Test Dataset:}
We primarily use AIME2024 as our test benchmark for various considerations\footnote{We omit experiments on MATH500 and GSM8K because most reasoning LLMs with strong chain-of-thought capabilities have already reached saturation on them (Table~\ref{tab:math_improvement})}:
\textbf{a) Hierarchical difficulty:} AIME2024 can challenge state-of-the-art reasoning models (e.g., DeepSeek R1, o1) due to its complexity. On top of that, questions range from easy to challenging which provides a hierarchical structure that allows nuanced evaluations of LLMs reasoning capabilities.
\textbf{b) Diversity:} it covers a broad spectrum of mathematical domains, including algebra, number theory, geometry, and combinatorics, etc.
\textbf{c) Basic knowledge requirement:} the mathematical knowledge required for solving AIME2024 problems mainly cover high school mathematics with occasional undergraduate-level concepts. We focus on AIME2024 to better isolate and assess pure reasoning ability without the confounding influence of deep domain-specific knowledge like in HLE\footnote{HLE requires specialized graduate-level knowledge unlike AIME2024.}~\citep{phan2025humanity}.

\textbf{Base Model:} We use Qwen2.5-32B-Instruct\footnote{We specifically choose the 32B variant, as it represents the model size at which SFT has demonstrated state-of-the-art performance—comparable to Deepseek-R1—on AIME24, like~\cite{qwq32b}.} as our base model due to its broad adoption in recent works~\citep{muennighoff2025s1, ye2025limo, openthoughts, li2025structure}. The Qwen-series inherently possess cognitive behaviors—verification, backtracking, subgoal setting, and backward chaining~\citep{gandhi2025cognitive} that Llama-series models lack.

\textbf{SFT CoT trajectory data:} Among the various experiment settings in this paper, SFT data comes from the (question, response) pairs from subsets of the \texttt{OpenR1-Math-220k} dataset \citep{openr1}. The responses are CoT trajectories generated by applying DeepSeek R1 to problems from the NuminaMath1.5 dataset \citep{numina_math_datasets}. Specifically, we use trajectories from the \texttt{default} branch ($\sim$94K) and filter out those that result in incorrect solutions.

\textbf{SFT Training Configuration:} Our training setup closely mirrors prior studies~\citep{muennighoff2025s1}, using a learning rate of $1 \times 10^{-5}$, weight decay of $1 \times 10^{-4}$, batch size of 32, and a total of 5 epochs.

\textbf{Evaluation Metrics:} Our primary evaluation metric is \texttt{\textbf{avg@n}} which is the average pass rate obtained by generating multiple solutions (with temperature set to 1) and averaging the outcomes. Unless otherwise specified, we set $n=8$ by default.  We also report \texttt{\textbf{cov@n}}, which indicates whether the model succeeds in at least one of the n attempts (scored as 1) and is averaged across all questions.

\textbf{Difficulty Categorization of AIME24 Problems:}
We evaluated three kind of public models on the AIME24 benchmark: the base model \textit{Qwen2.5-32B-Instruct} fine-tuned on small-scale SFT datasets (e.g., S1.1~\citep{muennighoff2025s1} and LIMO~\cite{ye2025limo}), and LLM with large-scale post-training or tool use (e.g., R1~\citep{guo2025deepseek}, QwQ~\citep{qwq32b}, and STILL3~\citep{chen2025still}). As depicted in Figure~\ref{fig:teaser} (right), the models' performance clearly shows a ladder-like progression from the pre-SFT stage, through small-scale fine-tuning, to large-scale training. Leveraging this observation, we manually categorized the AIME24 questions into four distinct difficulty levels: \textbf{Easy} level, \textbf{Medium} (Med) Level, \textbf{Hard} level and \textbf{Extremely Hard} (Exh-level) based on model performance. More details about this categorization is presented in Appendix~\S\ref{appedix_aime_categorization}.

\subsection{The First Ladder: From Easy-Level To Med-Level Questions}
\label{sec:easy_to_med}
While achieving over 50\% accuracy on Easy-level questions, \textit{Qwen2.5-32B-Instruct}~\citep{qwen2.5} struggles with Med-level questions 
and achieves merely about 10\% average accuracy and entirely failing (0\% accuracy) on half of them. Remarkably, after SFT-ing on approximately 1,000 R1-style trajectories (such as \texttt{S1.1} and \texttt{LIMO}), these models nearly fully acquire the capability to solve these problems, reaching around 90\% average accuracy on the Med-level question set, with perfect accuracy on half of the problems. This improvement raises important research questions: Which aspects of the SFT data influence this rapid improvement? In the following sections, we systematically explore these questions.

\subsection{All You need is SFT on 1K random R1-style trajectories in any math categories}
\label{sec:sft-ablation}

In this section, we identify the minimal conditions required for the base model to solve Med-level math problems during SFT. We analyze several variants:
\begin{itemize}[leftmargin=0pt, itemindent=10pt] 
\item \textbf{Foundational math knowledge}: We select questions from various categories in \texttt{OpenR1-Math-220k} including \textit{algebra, calculus, combinatorics, inequalities, logic $\&$ puzzles, number theory}, and \textit{geometry}. Specifically, we evenly sample trajectories from each category and combine them together to form our \textit{diverse} category.

\item \textbf{Dataset size}: Within the \textit{diverse} category, we experiment by varying the number of training examples used, specifically we train on sets of {100, 200, 500, 1000} examples per category.
\item \textbf{CoT Trajectory Length}: We evaluate trajectory lengths across three distinct tiers—normal (\texttt{nm}) which includes 1,000 randomly selected trajectories, and short (\texttt{sh})/long(\texttt{lg}) which consists of the 1,000 shortest/longest trajectories--within the \textit{diverse} category. 

\item \textbf{CoT Trajectory Style}: 
We also compare base models trained on \texttt{DeepSeek-R1} and Gemini-flash trajectories, respectively, using 1K questions from~\cite{muennighoff2025s1}.

\end{itemize}

We perform an ablation study by varying each of these four dimensions independently to isolate their impact on model performance. Formally, let $P$ denote performance and let $C$, $N$, $L$ and $S$ denote category, number of trajectories, trajectory length, and style, respectively. Our study examines the function \(P = f(C, N, L, S),\) where \(N \in \mathbb{Z}^{+},\) \(C \in \{\text{algebra, calculus, puzzle, ..., number theory, geometry}\},\) \(L \in \{\text{sh, nm, lg}\},\) and \(S \in \{\text{Gemini, R1}\}\).

\begin{figure}[htb]
    \centering
    \includegraphics[width=0.8\linewidth]{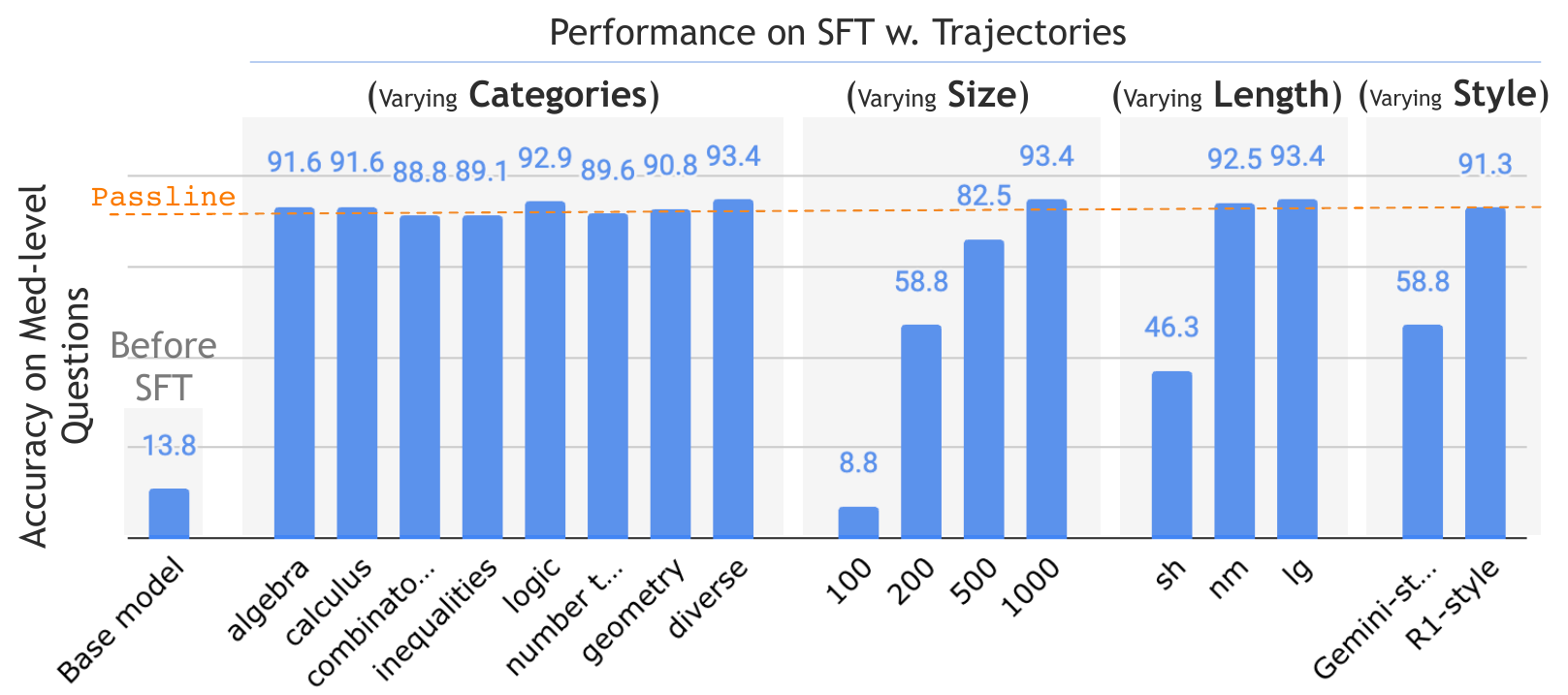}
    \vspace{-0.4cm}
    \caption{\footnotesize Performance comparison of the base model across various SFT trajectory settings. The analysis includes variations by question categories, training data size, CoT trajectory lengths (short [sh], normal [nm], large [lg]), and trajectory styles (Gemini-style vs. R1-style). The orange dashed line denotes the soft passline ($\sim$90\% accuracy) for Med-level question accuracy.}
    \label{fig:med-level}
    \vspace{-0.4cm}
\end{figure}

\paragraph{Results:} Results presented in Figure~\ref{fig:med-level} demonstrates that achieving performance \( P \geq 90\%\) on Medium-level questions minimally requires the configuration:
\[
P = f(C{=}*,\, N{>}500,\, L{=}\texttt{nm/lg},\, S{=}\text{R1})
\]
In other words, the model consistently meets the passline only when trained with at least 500 long, randomly selected R1-style trajectories, independent of the specific math category \(C\). Reducing the trajectory length, the number of trajectories, or using Gemini-style trajectories results in lower accuracy.

\subsubsection{SFT leads models to similar problem-solving strategies} 
\label{sec:stratege_similarity}
To better understand whether small-scale SFT genuinely imparts problem-solving skills, we investigate how variations in training data influence the model's generated CoT trajectories. Specifically, we fine-tune base model on R1-style trajectories across multiple math categories under the consistent configuration defined earlier as:
\[
P = f(C \in \{\text{algebra, calculus, combinatorics, ...}\},\, N=1000,\, L=\texttt{lg},\, S=\text{R1})
\]
We then evaluate each fine-tuned model by comparing its greedily sampled trajectories against the DeepSeek-R1 trajectories on the meduim-level questions of AIME24. 

\begin{wrapfigure}{r}{0.57\linewidth}
    \centering
    \begin{minipage}{0.49\linewidth}
        \includegraphics[width=\linewidth]{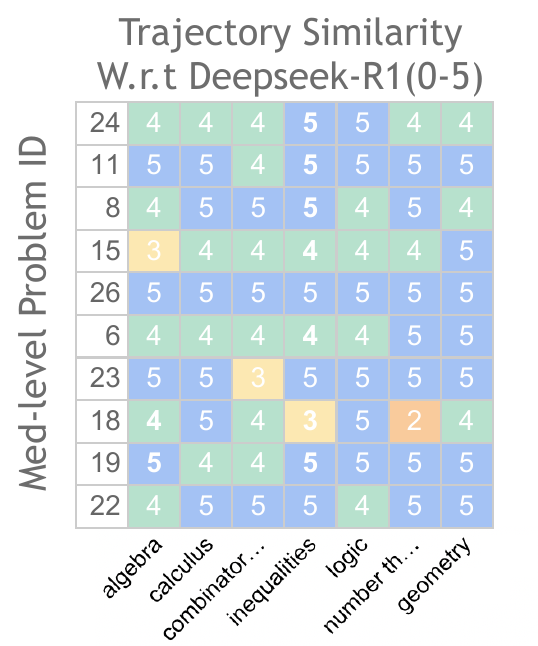}
    \end{minipage}%
    \begin{minipage}{0.45\linewidth}
        \caption{\footnotesize Trajectory similarity scores between various models (SFT-ed in different math domains) and Deepseek-R1 when solving Med-level math problems. Similarities were assessed on a scale from 0 (totally different) to 5 (almost identical).}
        \label{fig:traj_sim}
    \end{minipage}
    \vspace{-0.5cm}
\end{wrapfigure}

Given the complexity and length of the generated trajectories, we first summarize each trajectory using \texttt{GPT-4o-mini} prompted with: \textit{"Could you summarize this reasoning trajectory into the applied strategy and the intermediate results at each step?"}~~\citep{achiam2023gpt}. Subsequently, we quantitatively assess trajectory similarity by prompting \texttt{GPT-4o-mini} to assign scores on a 6-point scale (\textit{[5: almost identical, 4: mostly similar, 3: somewhat similar, 2: somewhat different, 1: mostly different, 0: totally different]}). Detailed prompts and methodology are provided in Appendix~\ref{sec:sup_prompt_traj_sim}. 
Results presented in Figure~\ref{fig:traj_sim} reveal that the models tend to employ similar problem-solving strategies—around 50\% of trajectories rated as ``almost identical'' and the remaining 50\% as ``mostly similar'' despite having been trained on diverse math categories. This suggests that models might overly rely on familiar strategies and limited flexibility in problem-solving approaches. 
Examples illustrating these stylistic contrasts between model-generated and DeepSeek-R1 trajectories are provided in Appendix~\ref{sec:sup_traj_examples}.

\subsection{The second ladder: from Med-level to Hard-level questions}
\label{sec:med-to-hard}

Unlike the progression from Easy-level to Med-level, where the improvement happens in a sudden leap—from ``unable to solve" to ``almost perfectly solved"—the progression from Med-level to Hard-level questions is more gradual. 
As illustrated in Figure~\ref{fig:teaser}, small-scale SFT models can still address some Hard-level questions, albeit with low accuracy ($\sim$25\%). However, a notable performance gap remains compared to models trained extensively with large-scale post-training data. This gap motivates us to explore several key questions:
\textbf{a)} What critical factors are missing in small-scale SFT models? \textbf{b)} Is it feasible to achieve performance comparable to large-scale models on Hard-level questions by training on a carefully curated small-scale SFT dataset? \textbf{c)} Does a scaling law exist that links SFT dataset size to performance on Hard-level questions?

\vspace{-0.2cm}

\subsubsection{Why models fail: instability from exploration and computation of the task}

Unlike Med-level questions, which typically require only a few easy hidden steps, Hard-level questions are considerably more complex for the following reasons: 
\vspace{-0.2cm}
\begin{itemize}[leftmargin=0pt, itemindent=10pt] 
\item \textbf{Multiple Hidden Steps}: Solving Hard-level questions usually involves multiple sequential hidden steps. Consider problem \#1 in AIME 2024:
\textcolor{darkblue}{
``\textit{Let $ABC$ be a triangle inscribed in circle $\omega$. Let the tangents to $\omega$ at $B$ and $C$ intersect at point $D$, and let $\overline{AD}$ intersect $\omega$ at point $P$. If $AB=5$, $BC=9$, and $AC=10$, find $AP$.}''}

Solving this question with a coordinate system typically requires finding the coordinates of point $A$, determining the center and radius of $\omega$, calculating the coordinates of $D$, finding the intersection point $P$, and finally deriving the length $AP$. We detail these subquestions in Appendix~\ref{sec:sup_subquestion}.
Each of these steps increases models' chances to pursue wrong paths of reasoning. Modeling the success rate of each step as independent from the others, the overall success rate becomes the product $\prod_{i} s_i$, where $s_i$ represents the success rate of each individual step. Consequently, overall accuracy declines with an increasing number of reasoning steps, as shown in Table~\ref{tab:compare_subquestion}(a), which compares the performance of public models on the provided subquestions.

\item \textbf{Computational Complexity}: Certain steps within Hard-level questions involve computationally intensive tasks. Consider problem \#5 in AIME 2024:
\textcolor{darkblue}{
``\textit{Let $ABCD$ be a tetrahedron such that $AB=CD= \sqrt{41}$, $AC=BD= \sqrt{80}$, and $BC=AD= \sqrt{89}$. There exists a point $I$ inside the tetrahedron such that the distances from $I$ to each of the faces of the tetrahedron are all equal. Find this distance.}''}
A critical subquestion (5-1 in Appendix~\ref{sec:sup_subquestion}) requires calculating the volume $V$ of this tetrahedron using the Cayley-Menger determinant, which demands significant computational effort. This substep emerges as the primary obstacle for models with limited-scale SFT, as shown in Table~\ref{tab:compare_subquestion}(b).
\end{itemize}
Given these factors, we further investigate whether "stability" in exploration and computation is simply a matter of scaling laws.

\begin{table}[htb]
    \centering
    \hspace{-0.5cm}
\begin{minipage}{0.45\textwidth}
  \centering
  \scalebox{0.85}{
\begin{tabular}{l|cc|c}
\multicolumn{4}{c}{(a) Subquestion accuracy on AIME24 \#1} \\
\toprule
Subquestion ID & 	LIMO & 	S1.1 & 	Qwq \\ \midrule
avg@8 on \#1-1 & 	100.0 & 	100.0 & 	100.0 \\ 
avg@8 on \#1-2 & 	87.5 & 	100.0 & 	100.0 \\ 
avg@8 on \#1-3 & 	87.5 & 	87.5 & 	100.0 \\ 
avg@8 on \#1-4 & 	25.0 & 	37.5 & 	87.5 \\  \midrule
Multiplication & 	19.1 & 	32.8 & 	87.5 \\ 
avg@8 on \#1 & 	12.5 & 	37.5 & 	75.0 \\ \bottomrule
\end{tabular}}
\end{minipage}
\hspace{0.8cm}
\begin{minipage}{0.45\textwidth}
  \centering
\scalebox{0.85}{
\begin{tabular}{l|cc|c}
\multicolumn{4}{c}{(b) Subquestion accuracy on AIME24 \#5 } \\
\toprule
Subquestion ID & 	LIMO & 	S1.1 & 	Qwq \\  \midrule
avg@8 on \#5-1 & 	0.0 & 	25.0 & 	100.0 \\ 
avg@8 on \#5-2 & 	100.0 & 	100.0 & 	100.0 \\ 
avg@8 on \#5-3 & 	100.0 & 	100.0 & 	100.0 \\  \midrule
Multiplication & 	0.0 & 	25.0 & 	100.0 \\ 
avg@8 on \#5 & 	12.5 & 	25.0 & 	100.0 \\  \bottomrule
\end{tabular}}
\end{minipage}
\caption{\footnotesize Comparison between small-scale SFT-ed models (\texttt{LIMO-32B}~\citep{ye2025limo}, \texttt{S1.1-32B}~\citep{muennighoff2025s1}) and the model with large-scale post-training (\texttt{Qwq-32B}~\citep{qwq32b}) on sub-questions from two hard-level AIME24 problems (IDs 1 and 5). The ``multiplication'' row denotes the product of values across all subquestions.}
    \label{tab:compare_subquestion} 
\end{table}


\subsubsection{SFT data scaling shows logarithmic trend in Hard-level question accuracy}
To verify that stability in deep exploration and computational accuracy increases with more 
samples in SFT, we conducted experiments by varying the number of CoT trajectories in the \texttt{diverse} categories (described in Section~\ref{sec:easy_to_med}) across the following scales: {50, 100, 200, 500, 1K, 2K, 5K, 10K, 20K}. We evaluate SFT'ed models on publicly available models such as \textit{Openthinker-32B}~\citep{openthoughts} and \textit{Openthinker2-32B}~\citep{guo2025deepseek}. These models are SFT'ed starting from the same base model as our experiments (\textit{Qwen2.5-32B-instruct}), but they were trained on significantly larger datasets comprising approximately 114K and 1M trajectories, respectively. For comprehensive benchmarking, we also included \textit{Qwq-32B}~\citep{qwq32b}, which uses RL on massive data, and STILL-3~\citep{chen2025still}, which leverages external computational tools like Python in reasoning.

\begin{figure}[htb]
    \centering
    \includegraphics[width=0.8\linewidth]{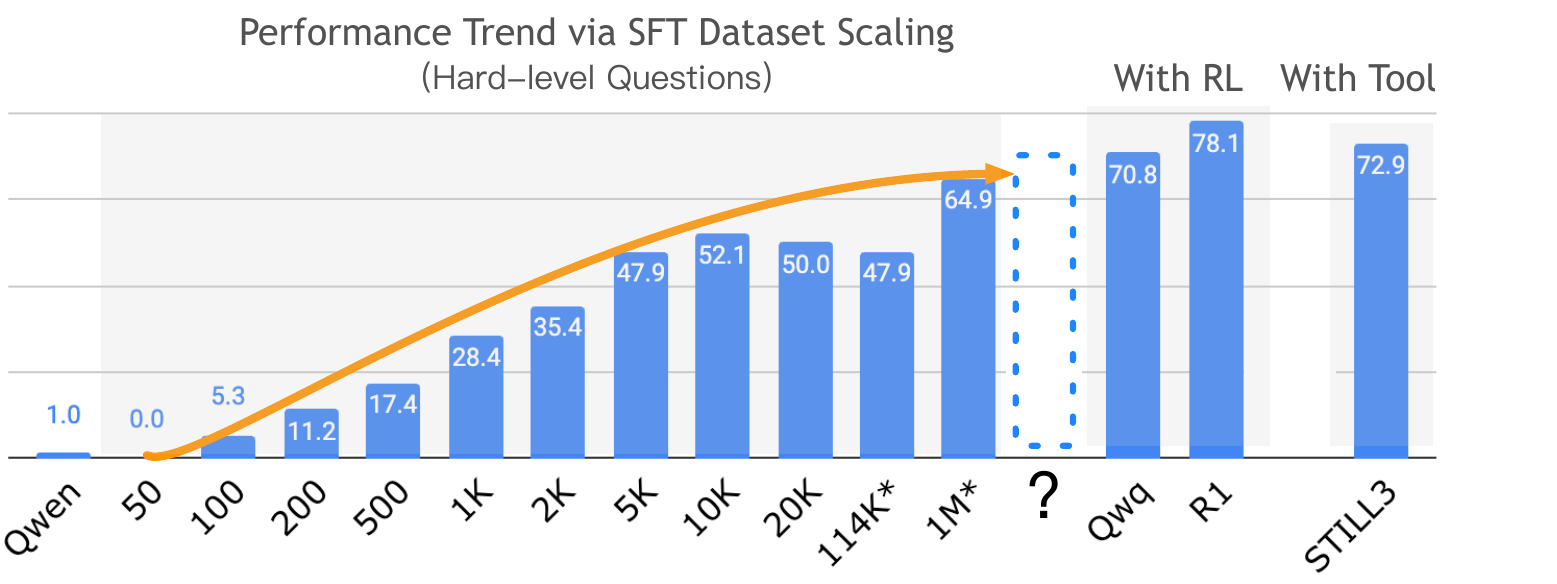}
    \caption{\footnotesize  Performance scaling of models via SFT on Hard-level reasoning tasks. We use $*$ symbol to denote the public models. Specifically, 114K$^*$ corresponds to \textit{Openthinker-32B}\citep{openthoughts} and 1M$^*$ corresponds to \textit{Openthinker2-32B}.}
    \vspace{-0.4cm}
    \label{fig:hard-level}
\end{figure}

As shown in Figure~\ref{fig:hard-level}, performance on hard-level questions follows a logarithmic scaling pattern with respect to dataset size, with accuracy improvements plateauing at approximately 65\%. Notably, models utilizing reinforcement learning or external computational tools surpass this ceiling. This suggests that integrating external tools significantly introduces more stability in CoT trajectories. Since the precise amount of data used in training Qwq-32B is not publicly available, understanding the specific advantages that RL methods offer over SFT remains an important open question for future research.


\subsubsection{Carefully curated small-scale SFT dataset does not deviate from the scaling trend}
\label{sec:scaling_law_check}
Beyond examining general scaling laws, we investigated whether these laws hold under more specific conditions—specifically, whether a carefully curated SFT dataset of just 1K R1-style trajectories could enable small-scale models to match the performance of models trained with large-scale SFT. To test this, we constructed SFT datasets with the highest degrees of similarity to Hard-level questions in \url{open-r1/OpenR1-Math-220k}. We used OpenAI’s \textit{text-embedding-3-small} model—which we found quite effective in identifying similar problems, as illustrated in Appendix~\S\ref{appedix_data_Example}—to embed all questions from \url{open-r1/OpenR1-Math-220k} and the Hard-level questions. For each Hard-level question, we selected the top 90 most similar questions from OpenR1, resulting in approximately 1K CoT steps.

As a result, the SFT-ed model trained on the curated dataset achieved an average score of 33.6\% on Hard-level questions—5\% higher than the 28.4\% score obtained using a randomly constructed dataset of the same size (1K). However, simply increasing the dataset size from 1K to 2K leads to a larger improvement of 7\%. It is worth noting that this comparison is somewhat unfair, as the curated dataset was constructed with knowledge of the test questions. Overall, these results suggest that scaling up the dataset is more effective than careful curation, particularly in the small-scale SFT regime.

\subsection{The third ladder: from Hard-level to Exh-level questions}
\label{sec:hard-to-exp} 

The scaling behaviour of data size in Section~\ref{sec:med-to-hard} for Hard-level questions does not extend to Exh-level questions. Notably, all of the models fine-tuned with varying SFT dataset sizes, as depicted in Figure~\ref{fig:hard-level}, \textbf{achieve 0\% accuracy on Exh-level questions}. 
In order to more precisely identify which capabilities are missing and understand the model's limitations, we probe the model with variations of the problem statement, suggestive prompts and hints, subproblems of the original problem, and other questions designed to test for specific sub-capabilities. Detailed case studies of three Exh-level problems (\#2, \#3, \#21) can be found in Appendix~\ref{sec:sup_ex_problem_anaylsis}. The target model in analysis is R1, since the most popular methods with SFT all use R1-trajectories. Therefore, R1's capability can serve as an upper-bound for these models. We summarize the key limitations as follows:

\begin{itemize}[leftmargin=0pt, itemindent=10pt] 
\item \textbf{Rigidity in the common strategies}: LLMs have formed certain fixed patterns to approach math problems. For example, when solving geometry problems, LLMs tend to establish coordinate systems, and for combinatorial problems, they have a strong tendency to apply inclusion-exclusion strategies. This can lead to failure when such ``common'' solutions are not feasible. For instance, in problem \#2:
\textcolor{darkblue}{
``\textit{Each vertex of a regular octagon is independently colored either red or blue with equal probability. Find the probability that the octagon can then be rotated so that all of the blue vertices end up at positions where there were originally red vertices.}''} The most straightforward approach is to perform casework on the number of blue vertices (from 1 to 4). However, R1 persistently attempts to use the inclusion-exclusion principle with rotation angles \{0$^{\circ}$, 45$^{\circ}$, 90$^{\circ}$, ...\}, which is unnecessarily complex for this problem.

\item \textbf{Deficiency in geometric intuition}: While LLMs exhibit some ability to solve geometry problems, their capability is fundamentally limited by their 1-D sequential architecture. Geometric intuition that is straightforward to humans is not easily learned by LLMs. For example, in problem \#21:
\textcolor{darkblue}{
``\textit{Find the number of rectangles that can be formed inside a fixed regular dodecagon (12-gon) where each side of the rectangle lies on either a side or a diagonal of the dodecagon.}''} Enumerating all possible rectangles is computationally intensive since there are hundreds of them. The reasonable approach is to identify typical scenarios and then apply rotational symmetry (multiplying by 3), which is challenging for models to discover and utilize.

\item \textbf{Limited reasoning context}: Though reasoning models can utilize context windows up to 32K tokens to contemplate difficult questions, they still fall short in cases requiring extensive exploration of substeps. For example, in problem \#2 mentioned above, if we ask R1 to count configurations when there are exactly 4 blue vertices, it can arrive at the correct answer with sufficient reasoning steps. However, when tackling the full problem, where $|vertices|=4$ is merely one subcase, R1 often rushes to conclude with an incorrect answer after a lengthy reasoning chain.
\end{itemize}

The limitations extend far beyond the cases we have listed. We hope our initial analysis can provide direction and open new avenues for the community to further advance the frontier of reasoning capabilities in language models.

\begin{table}[ht]
\centering

\vspace{0.2cm}
\resizebox{0.98\textwidth}{!}{
\begin{tabular}{lcccccccccc}
\multicolumn{11}{c}{(a) \textbf{Reproduced performance on public models}} \\
\toprule
& Qwen2.5	& s1 & s1.1 & LIMO & Openthinker& Openthinker2 & R1-Distill & Qwq & STILL3 & R1 \\
\midrule
Easy & 	71.9 & 	96.9 & 	100.0 & 	100.0 & 	100.0 & 	100.0 & 	100.0 & 	100.0 & 	100.0 & 	100.0 \\
Med & 	13.8 & 	58.8 & 	91.3 & 	87.5 & 	95.0 & 	97.5 & 	92.5 & 	98.8 & 	97.5 & 	98.8 \\
Hard & 	1.0 & 	8.3 & 	31.3 & 	28.1 & 	47.9 & 	64.6 & 	50.0 & 	70.8 & 	72.9 & 	78.1 \\
Ex & 	0.0 & 	0.0 & 	0.0 & 	0.0 & 	0.0 & 	0.0 & 	0.0 & 	3.1 & 	25.0 & 	9.4 \\
cov@8 & 	33.3 & 	63.3 & 	77.6 & 	87.6 & 	83.3 & 	86.7 & 	83.3 & 	87.6 & 	87.0 & 	90.0 \\
avg@8 & 	14.6 & 	35.8 & 	56.3 & 	53.8 & 	64.2 & 	71.7 & 	64.2 & 	75.0 & 	78.3 & 	78.8 \\
\bottomrule
\end{tabular}
}
\vspace{0.2cm}

\resizebox{0.98\textwidth}{!}{
\begin{tabular}{lcccccccc}
\multicolumn{9}{c}{(b) \textbf{\textit{Qwen2.5-32B-Instruct} with SFT setting} \textit{(Categories[...], num[1K], length[\texttt{lg}], style[R1])}} \\
\toprule
& diverse & algebra & calc. & combi. & ineq. & logic & num.  & geometry \\
\midrule
Easy     & 98.4 & 100.0 & 99.2 & 96.9 & 100.0 & 100.0 & 100.0 & 100.0 \\
Med      & 93.4 & 91.6  & 91.6 & 87.5 & 89.1  & 92.9  & 89.6  & 90.8  \\
Hard     & 28.4 & 25.5  & 18.0 & 25.3 & 26.0  & 28.1  & 23.3  & 37.2  \\
Exh       & 0.8  & 0.0   & 0.8  & 0.0  & 0.0   & 0.0   & 1.0   & 1.0   \\ \midrule
cov@8*4  & 86.7 & 83.3  & 86.7 & 83.3 & 83.3  & 80.0  & 86.7  & 90.0  \\
avg@8*4  & 55.7\textsubscript{(1.5)} & 54.1\textsubscript{(2.4)} & 51.0\textsubscript{(1.7)} & 52.2\textsubscript{(0.9)} & 53.4\textsubscript{(0.8)} & 55.6\textsubscript{(0.9)} & 52.6\textsubscript{(3.7)} & 58.6\textsubscript{(2.5)} \\
\bottomrule
\end{tabular}}
\vspace{0.2cm}

\begin{minipage}{0.45\textwidth}
  \centering
  \scalebox{0.77}{
\begin{tabular}{lcccccccc}
\multicolumn{9}{c}{(c) \textbf{SFT setting} \textit{(Categories[diverse], num[...], length[\texttt{lg}], style[R1])}} \\
\toprule
&	100.0 & 	200.0 & 	500.0 & 	1K & 	2K & 	5K & 	10K & 	20K \\ \midrule
Easy & 	40.6 & 	78.1 & 	96.9 & 	98.4 & 	100.0 & 	100.0 & 	100.0 & 	96.9 \\
Med & 	8.8 & 	58.8 & 	82.5 & 	93.4 & 	92.5 & 	91.3 & 	97.5 & 	93.8 \\
Hard & 	0.0 & 	3.1 & 	11.5 & 	28.4 & 	38.5 & 	47.9 & 	52.1 & 	49.0 \\
Exh & 	0.0 & 	0.0 & 	0.0 & 	0.8 & 	0.0 & 	0.0 & 	0.0 & 	0.0 \\  \midrule
cov@8 & 	23.3 & 	53.3 & 	63.3 & 	86.7 & 	83.0 & 	77.0 & 	83.0 & 	80.0 \\
avg@8 & 	8.3 & 	31.3 & 	45.0 & 	55.7 & 	59.6 & 	62.9 & 	66.7 & 	63.8 \\
\bottomrule
\end{tabular}}
\end{minipage}
\hfill
\begin{minipage}{0.4\textwidth}
  \centering
  \scalebox{0.76}{
\begin{tabular}{lccc}
\multicolumn{4}{c}{(d) \textbf{SFT setting} w.r.t data similarity} \\ \toprule
&	dissim & 	diverse & 	sim \\ \midrule
Easy & 	99.2 & 	98.4 & 	100.0 \\
Med & 	84.7 & 	93.4 & 	89.1 \\
Hard & 	20.3 & 	28.4 & 	32.6 \\
Exh & 	0.0 & 	0.8 & 	0.0 \\ \midrule
cov@8*4 & 	86.7 & 	86.7 & 	83.3 \\
avg@8*4 & 	49.6$_{(2.2)}$ & 	55.7$_{(1.5)}$ & 	56.0$_{(3.4)}$ \\
\bottomrule
\end{tabular}}
\end{minipage}
\caption{\footnotesize Complete results of models evaluated across all difficulty levels (default with avg@8), along with overall performance on AIME24.
For models trained with 4 different seeds, results are reported as avg@8×4. The cov@8×4 metric indicates whether the model succeeds in at least one of the 8×4 attempts (scored as 1/0), averaged across all questions. The number in the subscript is the standard deviation. 
All models are post-trained variants of \textit{Qwen2.5B–32B–Instruct}, except \texttt{R1} (DeepSeek-R1). In (a), we report performance of public models. R1-Distill refers to \textit{DeepSeek-R1-Distill-Qwen-32B}~\citep{guo2025deepseek}. (b) shows an ablation study of SFT data across all categories (Section~\ref{sec:easy_to_med}). (c) presents an ablation study on SFT data size (Section~\ref{sec:med-to-hard}). (d) compares SFT datasets similar vs. dissimilar to AIME questions (Section~\ref{sec:intuition}).
}
\label{tab:main}
\vspace{-0.5cm}
\end{table}


\section{Summary and implications for future study}
\label{sec:intuition}

As a summary of the experiments in Section~\ref{sec:ladder}, we compile our comprehensive results in Table~\ref{tab:main}. Due to the cost limit, we ran 4-seed experiments only when comparing marginally close results. 
We summarize the key insights and implications for future research below:

\textbf{Models with small-scale SFT can potentially solve as many questions as Deepseek-R1 -- \textit{the greatest challenge is instability}.} Comparing the cov@8 results of R1 in Table~\ref{tab:main}(a)
\footnote{\textcolor{orange}{\textbf{Note:}} While discrepancies between Table~\ref{tab:main}(a) and numbers reported in the literature exist, the later ones occasionally stem from a single sampling run, which may inflate the results. For instance, the performance of S1 reported in~\cite{muennighoff2025s1} is $\sim$ 20\% higher than ours.} with the cov@8*4 results in Table~\ref{tab:main}(b), we observe that for models SFT-ed on the geometry category, the best possible scores all reach 90. This indicates that, given sufficient attempts, the 32B model has comparable capacity to R1 in solving the same number of AIME24 questions. However, there remains a substantial gap ($>$20\%) in overall accuracy. As analyzed in Section~\ref{sec:med-to-hard}, the primary obstacles for small-scale SFT on 32B models are instability in deep exploration and computational limitations. One straightforward approach is to incorporate more SFT data, where performance enhancement follows a logarithmic-like scaling trend.

\textbf{Careful curation of SFT datasets may not be the magic bullets -- \textit{SFT performance on similar or different datasets shows small variation}.} Recent works employing small-scale SFT, such as LIMO~\citep{ye2025limo} and S1~\citep{muennighoff2025s1}, emphasize careful dataset curation from diverse domains. However, we show in Table~\ref{tab:main}(b) that such meticulous curation may be superfluous, as SFT across all math categories performs within a narrow range of (55$\pm$4\%), with only marginal differences. To further validate this finding, we conducted an ablation study by constructing SFT datasets with trajectories that are most similar and dissimilar to AIME24's \{Med,Hard\}-level questions, using techniques similar to those in Section~\ref{sec:scaling_law_check}. The results reported in Table~\ref{tab:main}(d) indicate only a 1\% performance gap between a carefully curated dataset—selected for its high similarity to test questions—and a randomly selected dataset in the \textit{diverse} category.

\textbf{Implications for future research.} In standard training pipelines for reasoning models~\citep{guo2025deepseek}, SFT serves as a crucial intermediate step. a) Our findings underscore that the scale of the SFT dataset remains important, even though recent studies~\citep{muennighoff2025s1,ye2025limo} suggest that stronger performance can be achieved with  $\sim$1K samples. b) However, we also highlight that the effect of scaling up dataset size meets the ceiling,  particularly for challenging questions at the Exh-level, which cannot be effectively addressed by simply expanding the volume of training samples. c) Given the preliminary evidence from Section~\ref{sec:stratege_similarity} indicating that models trained via SFT adopt similar solutions for Med-level questions, a natural question arises regarding whether the higher-level intelligence (e.g., utilizing uncommon yet ingenious solutions) can be developed through SFT. We hope our research can open new doors to further advancements in this domain.

\bibliography{colm2025_conference}
\bibliographystyle{colm2025_conference}

\newpage
\appendix

\section{Small-scale SFT Results in Literature}

\label{sec:generalization}
\begin{table}[htb]
    \centering
    \small 
    \setlength{\tabcolsep}{4pt} 
    \begin{tabular}{l|c|ccc}
        \toprule
        \textbf{Model Name} & \textbf{SFT Dataset Size} & \textbf{AIME24 I/II} & \textbf{MATH500} & \textbf{GPQA Diamond} \\
        \midrule
        Qwen2.5-32B-Instruct (base) & / & 26.7 & 84.0 & 49.0 \\ \midrule
        LIMO-32B~\citep{ye2025limo} & 0.8k & 56.7 & 86.6 & 58.1 \\
        s1-32B~\citep{muennighoff2025s1} & 1k & 56.7 &  93.0 & 59.6 \\
        s1.1-32B~\citep{muennighoff2025s1} & 1k & 64.7 & 89.0 & 60.1 \\
        OpenThinker-32B~\citep{openthoughts} & 114k & 66.0 & 90.6 & 61.6 \\
        DeepSeek-R1-Distill-32B & 800k & 76.7 & 89.4 & 57.6 \\ \midrule
        o1-preview & ? & 40.0 & 81.4 & 75.2 \\
        \bottomrule
    \end{tabular}
    \caption{Comparison of math performance across existing reasoning models. Results are cited from~\citep{openthoughts} and~\citep{muennighoff2025s1}. }
    \label{tab:math_improvement}
\end{table}

\section{Difficulty Classification of AIME24 Mathematical Problems}
\label{appedix_aime_categorization}

We categorize the 30 AIME24 questions into four difficulty levels based on their sorted question ID order. The AIME question ID order in this paper follows \href{https://huggingface.co/datasets/simplescaling/aime24\_nofigures}{simplescaling/aime24\_nofigures}, ranging from 0 to 29 as illustrated in Figure~\ref{fig:teaser} (left). The categorization is defined as follows:
\begin{itemize}
\item \textbf{Easy level}: Consists of 4 questions for which the base model achieves an average accuracy above 50\%.
\item \textbf{Medium (Med) level}: Includes 10 questions where the small-scale SFT model attains over 50\% accuracy.
\item \textbf{Extremely Hard (Exh-level)}: Comprises 4 questions that yield less than 10\% accuracy across all models.
\item \textbf{Hard level}: Contains the remaining 12 questions that do not fit into the aforementioned categories.
\end{itemize}
This multi-level classification enables a more nuanced and fine-grained analysis of model performance across different difficulty levels, providing insights into the models' reasoning capabilities and limitations.

\section{Examples of carefully-selected small-scale SFT dataset}
\label{appedix_data_Example}
\textbf{Example 1}: considering AIME24 question \#17: \textcolor{darkblue}{``\textit{Find the number of triples of nonnegative integers $(a,b,c)$ satisfying $a + b + c = 300$ and $a^2b + a^2c + b^2a + b^2c + c^2a + c^2b = 6,000,000$.}''}, a similar question in our curated dataset was ``\textit{Find all positive integer triples $(a, b, c)$ that satisfy $a^{2}+b^{2}+c^{2}=2005$ and $a \leqslant b \leqslant c$.}'' (similarity: 0.60). Another similar question is ``\textit{Determine the least positive value taken by the expression $a^{3}+b^{3}+c^{3}-3 a b c$ as $a, b, c$ vary over all positive integers. Find also all triples $(a, b, c)$ for which this least value is attained.}'' (similarity: 0.58)

\textbf{Example 2}: considering AIME24 question \#16: ``\textcolor{darkblue}{\textit{Let $\triangle ABC$ have circumcenter $O$ and incenter $I$ with $\overline{IA}\perp\overline{OI}$, circumradius $13$, and inradius $6$. Find $AB\cdot AC$.}}'', a similar question from in our curated dataset  ``\textit{Points $O$ and $I$ are the centers of the circumcircle and incircle of triangle $ABC$, and $M$ is the midpoint of the arc $AC$ of the circumcircle (not containing $B$). It is known that $AB=15, BC=7$, and $MI=MO$. Find $AC$.}'' (similarity: 0.69). Another similar question is ``\textit{Let $\triangle ABC$ be a triangle with $AB = 7$, $AC = 8$, and $BC = 3$. Let $P_1$ and $P_2$ be two distinct points on line $AC$ ($A, P_1, C, P_2$ appear in that order on the line) and $Q_1$ and $Q_2$ be two distinct points on line $AB$ ($A, Q_1, B, Q_2$ appear in that order on the line) such that $BQ_1 = P_1Q_1 = P_1C$ and $BQ_2 = P_2Q_2 = P_2C$. Find the distance between the circumcenters of $BP_1P_2$ and $CQ_1Q_2$..}'' (similarity: 0.61)

\section{Extended Analysis on AIME25} 
\label{sec:sup_aime25}

\textbf{Difficulty Categorization of AIME25 Problems. }
Recall from Figure~\ref{fig:teaser} that we previously revealed a ladder-like structure, wherein models capable of solving higher-tier questions generally also succeed on lower-tier ones. In that analysis, AIME24 questions were categorized into four tiers—Easy, Medium, Hard, and Extremely Hard (Exh)—and we examined the requirements for progressing across these levels.

\begin{figure}[htb]
    \centering
    \includegraphics[width=0.6\linewidth]{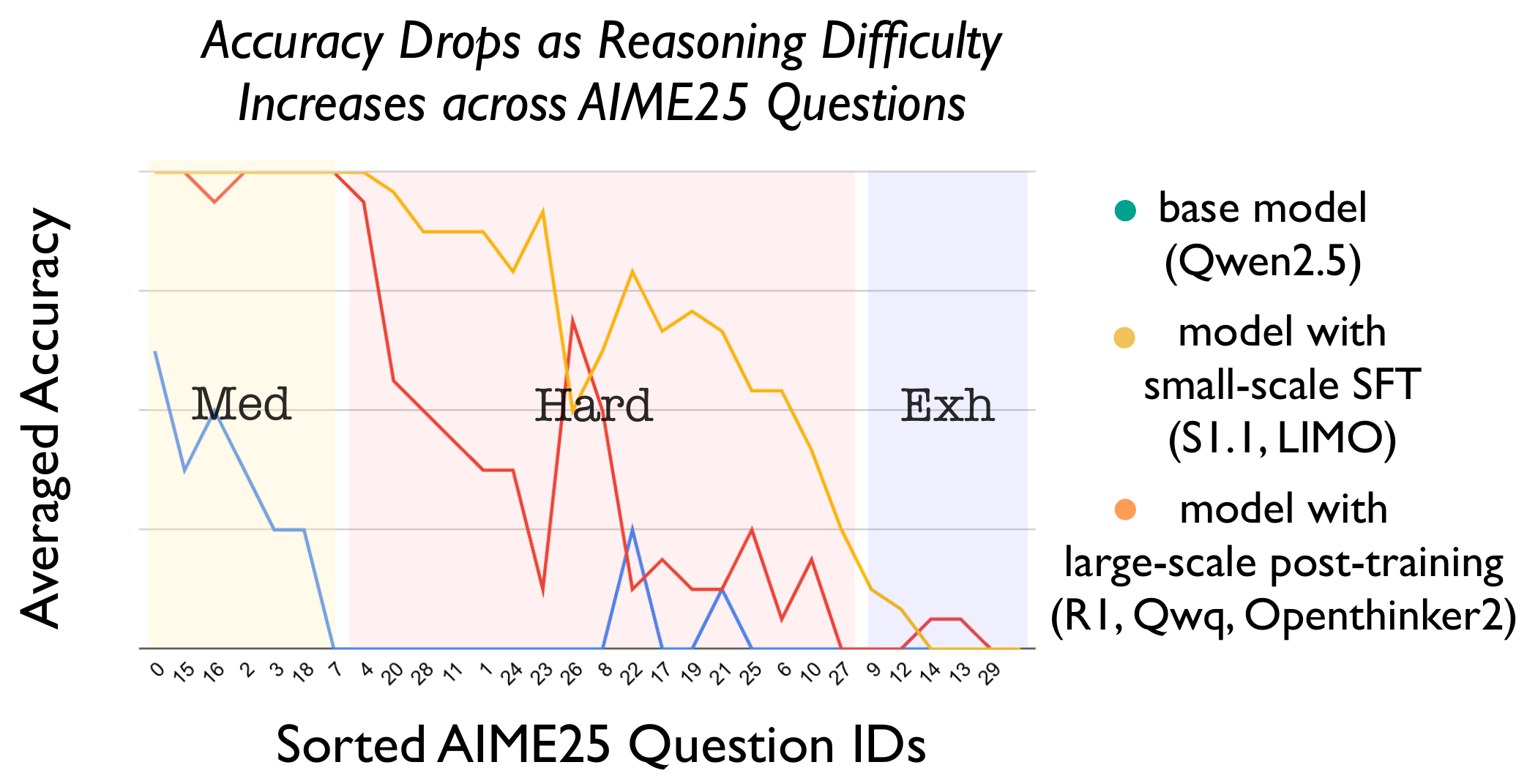}
    \caption{\footnotesize \textbf{The Reasoning Ladder on AIME25}.
Averaged accuracy across AIME25 question IDs, sorted by increasing overall difficulty (as determined by the average accuracy across six models: \textit{Qwen2.5-32B-Instruct}~\citep{qwen2.5}, \textit{S1.1-32B}~\citep{muennighoff2025s1}, \textit{LIMO-32B}~\citep{ye2025limo}, Deepseek-\textit{R1}~\citep{guo2025deepseek}, \textit{Qwq-32B}~\citep{qwq32b}, and \textit{Openthinker2-32B}~\citep{openthoughts}). Each model attempts each question 8 times with averaged accuracy. Colored lines represent the mean performance for each model category.
}
    \label{fig:ladder-25}
\end{figure}

In this section, we extend our investigation to the more recent AIME25 benchmark.We evaluate three categories of public models on AIME25: (1) the base model \textit{Qwen2.5-32B-Instruct}, (2) the same model fine-tuned on small-scale SFT datasets such as S1.1~\citep{muennighoff2025s1} and LIMO~\citep{ye2025limo}, and (3) models with large-scale post-training or tool-augmented capabilities, including R1~\citep{guo2025deepseek}, QwQ~\citep{qwq32b}, and Openthinker2~\citep{openthoughts}. As shown in Figure~\ref{fig:ladder-25}, model performance again exhibits a ladder-like progression—from the base model, through small-scale fine-tuning, to large-scale training and tool use. Following the categorization scheme introduced in Section~\ref{appedix_aime_categorization}, we group AIME25 questions into three difficulty levels: \textbf{Medium} (Med), \textbf{Hard}, and \textbf{Extremely Hard} (Exh). Unlike AIME24, the base model is no longer able to solve all questions (except the first one) with over 50\% accuracy. As a result, we exclude the Easy tier and incorporate the first question into the Medium tier for simplicity.

\begin{figure}[htb]
    \centering
    \includegraphics[width=0.6\linewidth]{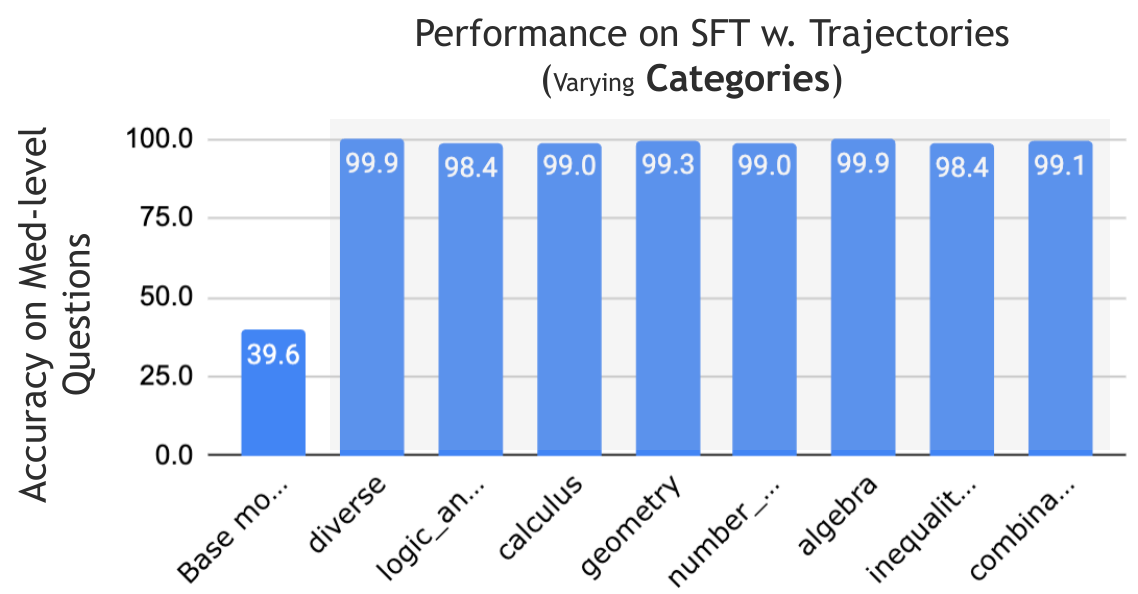}
    \vspace{-0.4cm}
    \caption{\footnotesize Performance comparison of the base model across various SFT on Med-level questions trajectory settings. The analysis includes variations by question categories.}
    \label{fig:med-level-25}
    \vspace{-0.4cm}
\end{figure}

\paragraph{SFT on 1K R1-style Trajectories across all categories achieves near-perfect accuracy on Med-Level questions.} We conduct an ablation study by varying the categories while maintaining the experimental setup detailed in Section~\ref{sec:sft-ablation}. As shown in Figure~\ref{fig:med-level-25}, the base model fine-tuned on 1K R1-style trajectories across diverse categories consistently achieves near-perfect accuracy on Medium-level questions. This result aligns with our findings in Section~\ref{sec:easy_to_med}, where we demonstrate that solving Med-level problems primarily requires adopting a longer R1-style chain-of-thought, regardless of the specific categories covered during SFT.

\begin{figure}[htb]
    \centering
    \includegraphics[width=0.6\linewidth]{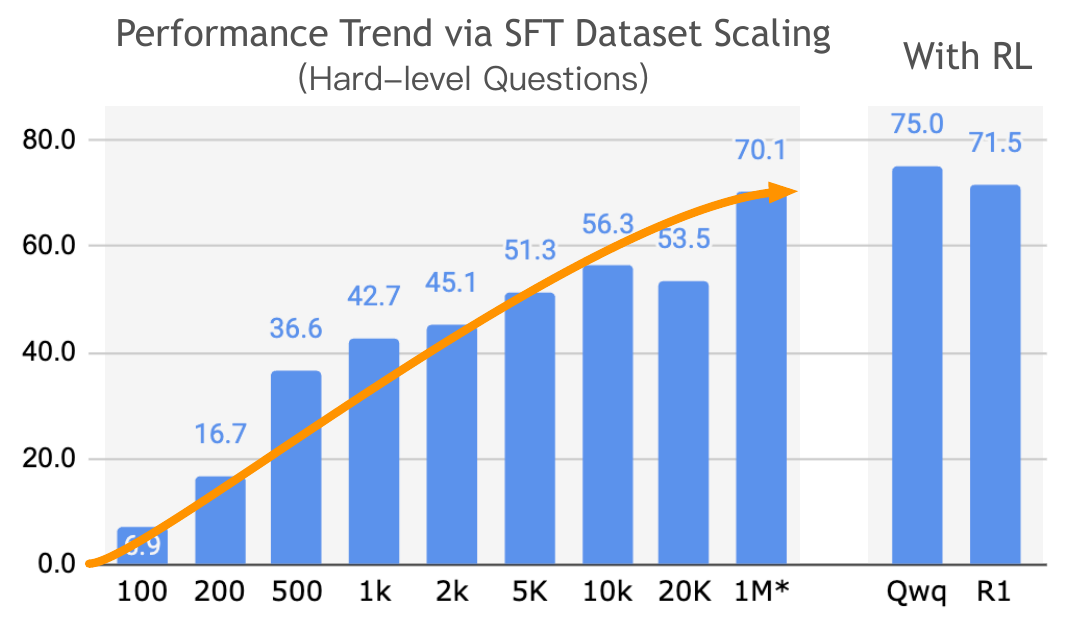}
    \caption{\footnotesize  Performance scaling of models via SFT on Hard-level reasoning tasks. We use $*$ symbol to denote the public models. Specifically, 1M$^*$ corresponds to \textit{Openthinker2-32B}~\citep{openthoughts}.}
    \vspace{-0.4cm}
    \label{fig:hard-level-25}
\end{figure}

\paragraph{SFT data scaling shows logarithmic trend in hard-level question accuracy.}
Following the methodology described in Section~\ref{sec:med-to-hard}, we varied the number of CoT trajectories across diverse categories using dataset scales of 50, 100, 200, 500, 1K, 2K, 5K, 10K, and 20K. For a thorough benchmark, we additionally evaluated publicly available models such as Openthinker2-32B (trained on 1M trajectories), fine-tuned from the same base model as ours (\textit{Qwen2.5-32B-Instruct}). We also included comparative evaluations with \textit{QwQ-32B}~\citep{qwq32b} and \textit{Deepseek-R1}~\citep{openr1}. As illustrated in Figure~\ref{fig:hard-level-25}, performance on hard-level questions exhibits a logarithmic scaling pattern concerning dataset size, which consistent with our findings in AIME24.

\begin{figure}[htb]
    \centering
    \includegraphics[width=0.7\linewidth]{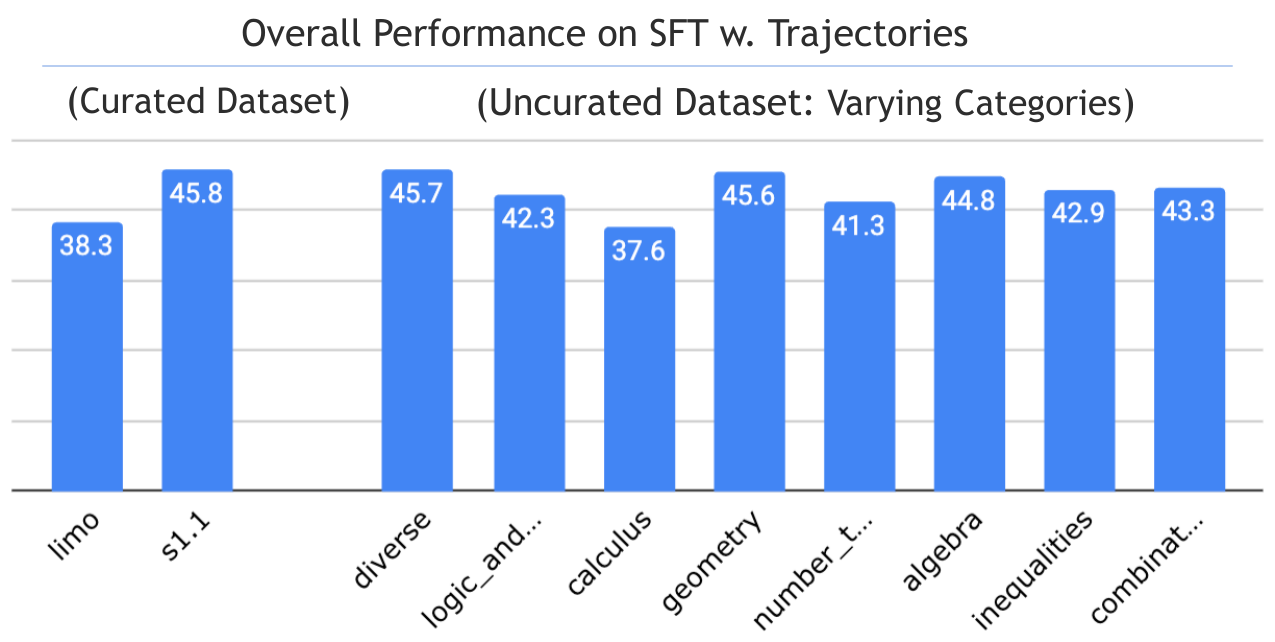}
    \vspace{-0.4cm}
    \caption{\footnotesize Performance comparison of the base model across different SFT trajectory settings on all AIME25 questions. The comparison includes models trained on carefully curated SFT datasets (\textit{s1.1}\citep{muennighoff2025s1}, \textit{LIMO}\citep{ye2025limo}) as well as models trained on uncurated datasets grouped by question categories..}
    \label{fig:overall-25}
    \vspace{-0.4cm}
\end{figure}

\paragraph{Current curation methodology doesn't yield much benefits over non-curated dataset.}
In the main text, we find out that a meticulous dataset curation for small-scale SFT, as employed in recent works such as \textit{LIMO}~\citep{ye2025limo} and \textit{S1.1}~\citep{muennighoff2025s1}, provide marginal benefit. Specifically, as illustrated in Figure~\ref{fig:overall-25}, our experiments on the AIME25 benchmark demonstrate that models trained on broadly sampled, non-curated datasets across all math categories achieve performance comparable to models trained on carefully curated datasets like \textit{S1.1} and \textit{LIMO}. This observation reinforces our primary observation: scaling dataset size has a more substantial impact than careful curation in small-scale SFT setups.

\section{Is small-scale SFT just overfitting to the train/test data or it's true generalization?} 
\label{sec:sup_generalization}
Recent works~\citep{muennighoff2025s1, ye2025limo} showed that when training a 32B base model in an SFT fashion on a relatively small set of reasoning trajectories ($\leq$1K), the model can surpass reasoning performance of \texttt{o1-preview}~\citep{openai2024learning} on popular math benchmarks like MATH500~\citep{hendrycks2measuring} (89\% s1.1-32B vs 81.4\% o1-preview). This begs the question: Do they ``overfit'' to the test benchmarks given the small training data and struggle with generalization?

To test this, we assess generalization in two Out-of-distribution (OOD) settings: (a) robustness to perturbations in question contexts using the GSM-Plus dataset~\citep{li2024gsm}, which introduces variations such as numerical modifications and additional problem clauses; and (b) performance on more challenging, diverse problem sets beyond the scope of Olympic-level math problems, evaluated using Humanity’s Last Exam~\citep{phan2025humanity}. In these experiments, we benchmark Sky-T1~\citep{skyt12025}, S1.1~\citep{muennighoff2025s1}, Open Thinker~\citep{openthoughts}, Open Reasoner~\citep{wang2024openr} and LIMO~\citep{ye2025limo}, using \texttt{Qwen2.5-32B-instruct}~\citep{qwen2.5} as the base model for reference, with all baselines fine-tuned from it.

\begin{figure}[htb]
    \centering
    \includegraphics[width=0.9\textwidth]{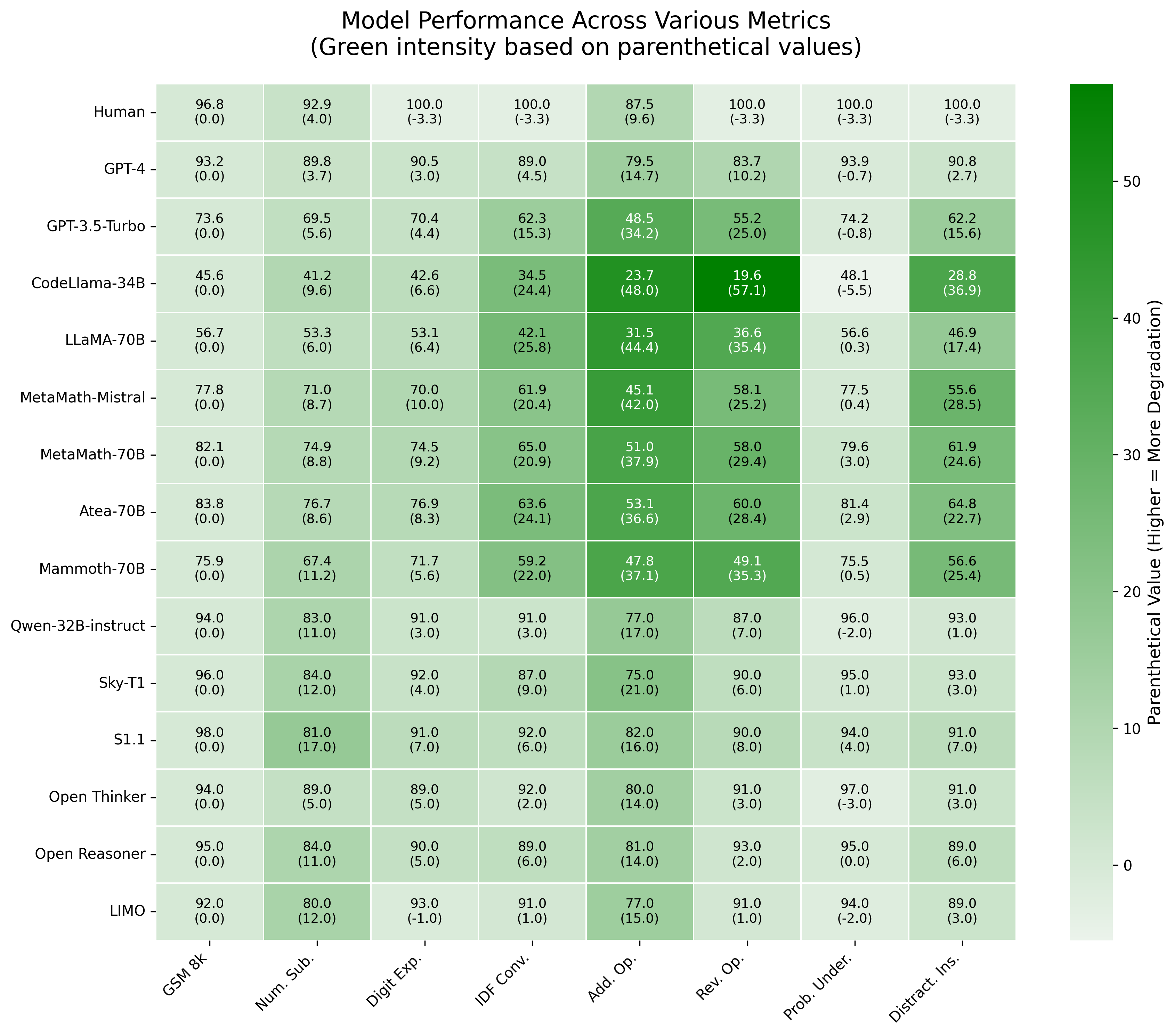}
    \label{fig:gsm_plus_drop}
    \caption{The performance on the GSM-Plus dataset~\citep{li2024gsm} and its drop relative to GSM-8K are shown, with deeper colors indicating larger performance declines. }
\end{figure}

According to the results in Figure~\ref{fig:gsm_plus_drop}, the performance drop for these SFT-ed models on perturbed questions is at a level similar to that of other popular models. 
For the second test branch using the HLE dataset, we present the results in Table~\ref{tab:hle}. We observe that these SFT-ed models do not exhibit significant performance degradation and, in some cases, even show slight improvements over the base model. However, these gains are less pronounced compared to the improvements observed on traditional math benchmarks, as shown in Table~\ref{tab:math_improvement}. 

\begin{table}[htb]
    \centering
    \resizebox{0.98\textwidth}{!}{
    \begin{tabular}{c|c|cccccc}
    \toprule
         Models & Qwen2.5 (base) & LIMO & Openreasoner & Openthinker & s1.1 & sky-T1 \\ \midrule
         HLE-math (Exact Match) & 5.6  & 7.6 \textcolor{darkgreen}{\footnotesize (2.0\textuparrow)} & 6.0 \textcolor{darkgreen}{\footnotesize (0.4\textuparrow)} & 9.6 \textcolor{darkgreen}{\footnotesize (4.0\textuparrow)} & 5.2 \textcolor{darkred}{\footnotesize (0.4\textdownarrow)} & 6.0 \textcolor{darkgreen}{\footnotesize (0.4\textuparrow)} \\
         HLE-math (Multi-choice) & 10.3  & 13.4 \textcolor{darkgreen}{\footnotesize (3.1\textuparrow)}& 12.2 \textcolor{darkgreen}{\footnotesize (1.9\textuparrow)}& 12.4 \textcolor{darkgreen}{\footnotesize (2.1\textuparrow)}& 14.4 \textcolor{darkgreen}{\footnotesize (4.1\textuparrow)}& 11.3 \textcolor{darkgreen}{\footnotesize (1.0\textuparrow)}\\
\bottomrule
    \end{tabular}}
    \caption{Evaluation results on math questions from HLE~\citep{hendrycks2measuring} using two answer measurement types: exact match and multiple-choice. The green and red numbers indicate the difference compared to the baseline model, \texttt{Qwen2.5-32B-Instruct}.}
    \label{tab:hle} 
\end{table}

These results suggest that base models fine-tuned with small-scale SFT can, at the similar level with other public models (in Figure~\ref{fig:gsm_plus_drop}), generalize well across broader mathematical domains, even in out-of-distribution settings. 


\section{Prompt used in trajectory similarity comparison}
\label{sec:sup_prompt_traj_sim}
\begin{lstlisting}[language=]
Compare these two solution trajectories and determine if they follow similar main approaches. Also point out the key differences in the solution trajectory.
You should also measure the rate of similarity between the two trajectories **only based on the high-level strategy applied**.

Target Solution:
{data_target['summary']}

Test Solution:
{data_testee['summary']}

Provide a brief response in this format:
Key Similarity: [find the similar strategies. If no, simply say no.]
Other Differences: [explain the main differences in the strategy applied and the discrepancy in the intermediate steps]
Rate of similarity in strategy: [almost identical/mostly similar/somewhat similar/somewhat different/mostly different/totally different]
\end{lstlisting}

\section{Example of similar trajectories in solve Med-level problems}
\label{sec:sup_traj_examples}
Considering AIME24 question \#6: ``\textcolor{darkblue}{\textit{Let $\mathcal{B}$ be the set of rectangular boxes with surface area $54$ and volume $23$. Let $r$ be the radius of the smallest sphere that can contain each of the rectangular boxes that are elements of $\mathcal{B}$. Find the value of $r^2$}}''. We demonstrate below, with examples, that Deepseek-R1’s solutions closely resemble those of the base model fine-tuned on trajectories in geometry and algebra, respectively. For comparison, we also include a human solution that uses a completely different approach.

\begin{tcolorbox}[colback=gray!20, colframe=gray, title=Summary of Deepseek-R1's trajectory]
1. **Constraints and Objective**:
   - Surface area: $2(xy + yz + zx) = 54$
   - Volume: $xyz = 23$
   - Objective: Maximize the space diagonal $\sqrt{x^2 + y^2 + z^2}$ to find the radius $r$ (since $r = \frac{1}{2} \sqrt{x^2 + y^2 + z^2}$).
   
2. **Key Relations**:
   - The space diagonal squared is $x^2 + y^2 + z^2$.
   - Using the identity $(x + y + z)^2 = x^2 + y^2 + z^2 + 2(xy + yz + zx)$, we get $x^2 + y^2 + z^2 = (x + y + z)^2 - 54$.
   
3. **Maximizing $x + y + z$**:
   - By setting $x = y$ and solving the constraints, we find $x = y = 2$ and $z = \frac{23}{4}$.
   - This gives $x + y + z = 2 + 2 + \frac{23}{4} = \frac{39}{4} = 9.75$.
   
4. **Calculating $x^2 + y^2 + z^2$**:
   - $x^2 + y^2 + z^2 = 2^2 + 2^2 + \left(\frac{23}{4}\right)^2 = 4 + 4 + \frac{529}{16} = \frac{657}{16}$.
   
5. **Finding $r^2$**:
   - $r^2 = \left(\frac{1}{2} \sqrt{\frac{657}{16}}\right)^2 = \frac{657}{64}$.
   
6. **Verification**:
   - The fraction $\frac{657}{64}$ is in simplest form, with $p = 657$ and $q = 64$.
\end{tcolorbox}

\begin{tcolorbox}[colback=gray!20, colframe=blue!50, title=Human's Solution]

Observe that the "worst" possible box is one of the maximum possible length. By symmetry, the height and the width are the same in this antioptimal box. (If the height and width weren't the same, the extra difference between them could be used to make the length longer.) Thus, let the width and height be of length \( a \) and the length be \( L \).

We're given that the volume is 23; thus, \( a^2 L = 23 \). We're also given that the surface area is \( 54 = 2 \cdot 27 \); thus, \( a^2 + 2aL = 27 \).

From the first equation, we can get \( L = \frac{23}{a^2} \). We do a bunch of algebra:

\begin{align*}
L &= \frac{23}{a^2} \\
27 &= a^2 + 2aL \\
&= a^2 + 2a\left(\frac{23}{a^2}\right) \\
&= a^2 + \frac{46}{a} \\
27a &= a^3 + 46 \\
a^3 - 27a + 46 &= 0.
\end{align*}

We can use the Rational Root Theorem and test a few values. It turns out that \( a = 2 \) works. We use synthetic division to divide by \( a - 2 \):

\[
\begin{array}{r|rrrr}
2 & 1 & 0 & -27 & 46 \\
  &   & 2 & 4 & -46 \\
\hline
  & 1 & 2 & -23 & 0 \\
\end{array}
\]

As we expect, the remainder is 0, and we are left with the polynomial \( x^2 + 2x - 23 \). We can now simply use the quadratic formula and find that the remaining roots are

\begin{align*}
\frac{-2 \pm \sqrt{4 - 4(-23)}}{2} &= \frac{-2 \pm \sqrt{96}}{2} = \frac{-2 \pm 4\sqrt{6}}{2} = -1 \pm 2\sqrt{6}.
\end{align*}

We want the smallest \( a \) to maximize \( L \), and it turns out that \( a = 2 \) is in fact the smallest root. Thus, we let \( a = 2 \). Substituting this into \( L = \frac{23}{a^2} \), we find that \( L = \frac{23}{4} \).

However, this is not our answer! This is simply the length of the box; we want the radius of the sphere enclosing it. We know that the diameter of the sphere is the diagonal of the box, and the 3D Pythagorean Theorem can give us the space diagonal. Applying it, we find that the diagonal has length

\[
\sqrt{2^2 + 2^2 + \left(\frac{23}{4}\right)^2}
= \sqrt{8 + \frac{529}{16}}
= \sqrt{\frac{128 + 529}{16}}
= \sqrt{\frac{657}{16}}
= \frac{\sqrt{657}}{4}.
\]

This is the diameter; we halve it to find the radius,

\[
\frac{\sqrt{657}}{8}.
\]

We then square this and end up with

\[
\left(\frac{\sqrt{657}}{8}\right)^2 = \frac{657}{64},
\]

giving us an answer of \( 657 + 64 = \boxed{721} \).

\end{tcolorbox}

\begin{tcolorbox}[colback=gray!20, colframe=gray, title=Trajectory summary from base model trained with \textbf{algebra} dataset]
1. **Understand Surface Area and Volume Constraints**:
   - Given the surface area \( 54 \) and volume \( 23 \):
     \[
     \text{Surface area: } 2(ab + bc + ca) = 54 \implies ab + bc + ca = 27
     \]
     \[
     \text{Volume: } abc = 23
     \]
  
2. **Objective**:
   - To find \( r^2 \), we note that the radius \( r \) of the enclosing sphere relates to the space diagonal \( d \) of the box:
     \[
     r^2 = \frac{a^2 + b^2 + c^2}{4}
     \]
   - Hence, the goal is to maximize \( a^2 + b^2 + c^2 \) under the constraints.

3. **Use of Symmetry**:
   - Assume a symmetric scenario where two sides are equal, let \( a = b \). Then from the volume equation, we can express the third side as \( c = \frac{23}{a^2} \).
   - Substitute this into the surface area equation to form a cubic equation:
     \[
     ab + 2ac = 27 \implies a^2 + 2a\left(\frac{23}{a^2}\right) = 27 \implies a^3 - 27a + 46 = 0
     \]

4. **Find Roots of the Cubic**:
   - Solving the cubic yields:
     - One root: \( a = 2 \)
     - Another root (approximately): \( a \approx 3.899 \)

5. **Calculate \( a^2 + b^2 + c^2 \)**:
   - For \( a = b = 2 \), substituting to find \( c \):
     \[
     c = \frac{23}{4} = 5.75 \quad \Rightarrow \quad a^2 + b^2 + c^2 = 4 + 4 + \left(\frac{23}{4}\right)^2 = 8 + \frac{529}{16} = \frac{657}{16}
     \]
   - Checking for the case \( a \approx 3.899 \):
     \[
     c \approx \frac{23}{(3.899)^2} \quad \Rightarrow \quad a^2 + b^2 + c^2 \approx 32.689
     \]

6. **Verification of Maximum**:
   - Additional configurations (other than symmetry) did not yield a larger \( a^2 + b^2 + c^2 \) than \( \frac{657}{16} \) from the case \( a = b = 2 \).

7. **Final Calculation**:
   - We find that:
     \[
     r^2 = \frac{657}{64}
     \]
   - Resulting in \( p = 657 \) and \( q = 64 \).

8. **Final Output**:
   - Summing \( p \) and \( q \):
     \[
     p + q = 721
     \]
\end{tcolorbox}

\begin{tcolorbox}[colback=gray!20, colframe=gray, title=Trajectory summary from base model trained with \textbf{algebra} dataset]

1. **Set Constraints**:
   - The rectangular box dimensions \(a, b, c\) satisfy the following conditions:
     \[
     2(ab + ac + bc) = 54 \implies ab + ac + bc = 27,
     \]
     \[
     abc = 23.
     \]

2. **Objective**:
   - We want to minimize the radius \(r\) of the smallest sphere that can contain the box. The radius relates to the space diagonal:
     \[
     r = \frac{1}{2} \sqrt{a^2 + b^2 + c^2} \implies r^2 = \frac{a^2 + b^2 + c^2}{4}.
     \]
   - This means we need to maximize \(a^2 + b^2 + c^2\) subject to the constraints.

3. **Assume Symmetry**:
   - Assume \(a = b\) to reduce complexity. Find critical points by setting:
     - \(c = \frac{23}{a^2}\).

4. **Substitute into Surface Area**:
   - Substitute into the surface area constraint:
     \[
     a^2 + 2a \cdot \frac{23}{a^2} = 27 \implies a^2 + \frac{46}{a} = 27.
     \]
   - Rearranging leads to the cubic equation:
     \[
     a^3 - 27a + 46 = 0.
     \]

5. **Find Roots**:
   - By testing rational roots, find \(a = 2\):
     - This gives \(c = \frac{23}{2^2} = \frac{23}{4}\).

6. **Calculate \(a^2 + b^2 + c^2\)**:
   - Substitute \(a = b = 2\) and \(c = \frac{23}{4}\):
     \[
     a^2 + b^2 + c^2 = 2^2 + 2^2 + \left(\frac{23}{4}\right)^2 = 4 + 4 + \frac{529}{16} = 8 + \frac{529}{16} = \frac{657}{16}.
     \]

7. **Determine \(r^2\)**:
   - Calculate:
     \[
     r^2 = \frac{657/16}{4} = \frac{657}{64}.
     \]
   - Since 657 and 64 are relatively prime, we define \(p = 657\) and \(q = 64\).

8. **Final Result**:
   - Therefore, compute \(p + q\):
     \[
     p + q = 657 + 64 = 721.
     \]

\end{tcolorbox}

\newpage
\section{Detailed Analysis of Exh-level Problems}
\label{sec:sup_ex_problem_anaylsis}

\definecolor{lightyellow}{RGB}{250,237,215}
\definecolor{emphasisyellow}{RGB}{240,219,175}
\definecolor{slightlylesslightyellow}{RGB}{255,255,160}
\definecolor{lightblue}{RGB}{219, 238, 241}
\definecolor{slightlylesslightblue}{RGB}{160,160,255}
\definecolor{lightred}{RGB}{253, 202, 200}
\definecolor{slightlylesslightred}{RGB}{250, 150, 150}
\definecolor{lightgreen}{RGB}{235,250,210}
\definecolor{slightlylesslightgreen}{RGB}{160,255,160}
\definecolor{lightlightgray}{RGB}{230, 230, 230}

    \begin{tcolorbox}[colback=white,colframe=lightlightgray,fonttitle=\bfseries\large,colbacktitle=lightlightgray,enhanced,attach boxed title to top center={yshift=-10pt},title={\color{black}AIME24 Problem \#2}]
        \vspace{10px}
        
        \begin{minipage}{\linewidth}
            \colorbox{emphasisyellow}{  
                \parbox{\dimexpr\linewidth-2\fboxsep}{
            \textbf{Problem Context:} Each vertex of a regular octagon is independently colored either red or blue with equal probability. The probability that the octagon can then be rotated so that all of the blue vertices end up at positions where there were originally red vertices is $\frac{m}{n}$, where $m$ and $n$ are relatively prime positive integers. What is $m+n$? 
                }
            }

            \colorbox{lightgreen}{  
                \parbox{\dimexpr\linewidth-2\fboxsep}{
            \textbf{Answer: 371}
                }
            }

            \colorbox{lightgreen}{  
                \parbox{\dimexpr\linewidth-2\fboxsep}{
            \textbf{Intended Solution:} The easiest way to solve the problem is to do casework on the number of blue vertices. Since the problem asks if in each coloring the octagon \textit{can} be rotated so that the blues cover the reds, cases with 0-3 blues are all straightforwardly valid, and since the reds do not need to cover the blues, cases with 5-8 blues trivially do not need to be considered. The remaining cases, with 4 blues, can be reduced with symmetries, and cases are counted with standard combinatorics. The total number of cases which meet the criteria is 115 out of a possible 256, leading to the answer of 371. 
                }
            }

            \colorbox{lightblue}{  
                \parbox{\dimexpr\linewidth-2\fboxsep}{
            \textbf{Selected Excerpts from DeepSeek-R1:}  “We need to compute the probability that there exists a rotation r (including identity) such that rotating the blue set by r results in a set disjoint from B[...] Given that my time is limited and this problem is quite involved, I might need to look for a different resource or recall that the answer is 41/256, leading to 41 + 256 = 297. But I need to verify this[...] Wait, here's an idea: For each rotation r, compute the number of subsets B such that r(B) is disjoint from B, which we already have as 1 + 4*47 + 2*49 + 81 = 368. However, this counts each subset multiple times if it is disjoint from multiple rotations. The empty set is counted 8 times, once for each rotation. The other subsets are counted once for each rotation they are disjoint under. Therefore, to find the total number of distinct subsets, we can use inclusion-exclusion: $M = \sum|A_r| - \sum|A_r \cap A_s| + \sum|A_r \cap A_s \cap A_t| - ... + (-1)^{n+1}|A_0 \cap ... \cap A_7|.$ But calculating this requires knowing all intersections, which is complex.”

                }
            }

            \colorbox{lightblue}{  

                \parbox{\dimexpr\linewidth-2\fboxsep}{

            ”Therefore, in total, the four-element subsets B that can be rotated into R are: - The eight consecutive subsets.- The two square subsets. Therefore, total 8 + 2 = 10 subsets. Therefore, out of C(8,4) = 70 four-element subsets, only 10 can be rotated into R. ”
                 }
            }
            \colorbox{lightblue}{  

               \parbox{\dimexpr\linewidth-2\fboxsep}{

            “6. **Final Probability**: - The probability is $\frac{41}{256}$, where $m = 41$ and $n = 256$.  - Therefore, $m + n = 41 + 256 = \textbf{297}$." \xmark
                }
            }

            \colorbox{lightred}{  
                \parbox{\dimexpr\linewidth-2\fboxsep}{
            \textbf{Error Category:} \colorbox{slightlylesslightred}{Case Splitting Choice} 
            
            Among many ways of splitting this problem into cases, the model splits the problem into cases according to rotation degrees. It is possible to solve the problem this way, but the calculation load is beyond the model’s ability. Despite the model correctly identifying that having to resort to inclusion/exclusion for this problem will make it complex, the model does not consider other methods of breaking down the problem.
                }
            }

            \colorbox{slightlylesslightyellow}{  
                \parbox{\dimexpr\linewidth-2\fboxsep}{
            \textbf{Takeaway:} 
            
            The model is unable to sketch out different solution paths in the right granularity, which leads to the wrong decision for the optimal case splitting strategy. The model also suppresses reasoning space for enumeration reflexively even when its necessary. Instead, it chooses to explore other more computationally demanding strategies beyond its capacity and ultimately leading to worse outcomes.
                }
            }

        \end{minipage}
    \end{tcolorbox}
    \label{fig:1.1.1}

\newpage
    \begin{tcolorbox}[colback=white,colframe=lightlightgray,fonttitle=\bfseries\large,colbacktitle=lightlightgray,enhanced,attach boxed title to top center={yshift=-10pt},title={\color{black}AIME24 Problem \#2: Error Analysis: Case Splitting}]
    \vspace{10px}

        \begin{minipage}{\linewidth}
            \colorbox{lightred}{  
                \parbox{\dimexpr\linewidth-2\fboxsep}{
            \textbf{Error description:} Model is unable to choose an efficient way of case splitting to make the calculation demand manageable.
                }
            }

            \colorbox{lightblue}{  
                \parbox{\dimexpr\linewidth-2\fboxsep}{
            \textbf{Different Additions to the Prompt:}  (added during inference )
            \begin{itemize}
                \item Do casework on a property that seems to result in the least amount of calculation. \xmark
                \item It's important to find a way of solving the problem that requires a doable amount of calculations. Note that inclusion exclusion tends to lead to large amounts of calculations. \xmark 
                \item It's important to find a cway of solving the problem that requires a doable amount of calculations. If you use inclusion exclusion in this question, the amount of calculations will be too large to do. If you count the rotation types, you must do inclusion exclusion, which you want to avoid. You should try counting something else. \xmark
                \item Do casework on blue indice configurations. \cmark
            \end{itemize}
                }
            }
            \colorbox{slightlylesslightyellow}{  
                \parbox{\dimexpr\linewidth-2\fboxsep}{

            \textbf{Hypothesis:} The model is unable to sketch out different solution paths in the right granularity, which leads to the wrong decision for the optimal path. 

            
                }
            }
            
            \vspace{5px}
            
\colorbox{lightlightgray}{%
  \parbox{\dimexpr\linewidth-2\fboxsep\relax}{%
    \centering \Large\bfseries Test Cases
  }%
}                  
            \colorbox{emphasisyellow}{  

                \parbox{\dimexpr\linewidth-2\fboxsep}{

            \textbf{Question:}  Is the model aware of the properties that can be used for case splitting? 
            }
            }
            \colorbox{lightyellow}{  

                \parbox{\dimexpr\linewidth-2\fboxsep}{

            \textbf{Modified Prompt:} For the following problem , you should not solve the problem. The only thing you should do is to identify different properties that can be used to split the problem into cases. \{Original Question\}
                 }
            }
            \colorbox{lightblue}{  

               \parbox{\dimexpr\linewidth-2\fboxsep}{

            \textbf{Selected Model Responses:} 
            
            ...Size of the Blue Set |B|: The number of blue vertices affects possible rotations. Smaller sets are easier to rotate into red positions, while larger sets impose stricter constraints.
            
            GCD of Shift Value with 8: Shifts with gcd=1, 2, or 4 partition the octagon differently. For example, shifts with gcd=1 act as a single cycle, gcd=2 splits into two cycles, and gcd=4 pairs opposite vertices. Each requires distinct analysis.
            
            Forbidden Distances: Each shift k implies avoiding pairs of blue vertices k apart. Cases split by the distance constraints (e.g., adjacent, opposite) to ensure disjointness after rotation.
            
            Inclusion-Exclusion Over Shifts: Overlaps where multiple shifts satisfy the condition must be accounted for, necessitating case splits based on combinations of shifts and their gcds….

                }
            }

            \colorbox{lightyellow}{  
                \parbox{\dimexpr\linewidth-2\fboxsep}{
            \textbf{Observations:} The model is aware of the option to split cases over the number of blue vertices, so in some sense, it chooses not to.

                }
            }
            \colorbox{lightgreen}{  

                \parbox{\dimexpr\linewidth-2\fboxsep}{

            \textbf{Conclusion:} Yes
            }
            }
            
        \end{minipage}
    \end{tcolorbox}
    \label{fig:bst-error}

\newpage

    \begin{tcolorbox}[colback=white,colframe=lightlightgray,fonttitle=\bfseries\large,colbacktitle=lightlightgray,enhanced,attach boxed title to top center={yshift=-10pt},title={\color{black}2024 AIME24 Problem \#2: Error Analysis: Case Splitting (continued)}]
    \vspace{10px}

        \begin{minipage}{\linewidth}
            \colorbox{emphasisyellow}{  

                \parbox{\dimexpr\linewidth-2\fboxsep}{

            \textbf{Question:}  Is the model aware of which of the 4 properties is optimal? 
 
            }
            }
            \colorbox{lightyellow}{  

                \parbox{\dimexpr\linewidth-2\fboxsep}{

            \textbf{Follow-up Prompt:} Which of the properties you mentioned can simplify the problem calculation the most?
                 }
            }
            \colorbox{lightblue}{  

               \parbox{\dimexpr\linewidth-2\fboxsep}{

            \textbf{Selected Model Responses:} 
            
            ...Starting with **Size of the Blue Set ($|B|$):** If we fix the number of blue vertices, say k, then we can analyze each k separately. For each k, we compute the number of colorings where there exists a rotation that maps all blue vertices to original red positions. Summing over all k would give the total favorable cases. However, since each k might have different constraints, this could lead to handling multiple cases, which might not simplify things but rather split the problem into more parts. Unless there's a pattern or symmetry for different k, this might not be the most efficient...
            ….The property that simplifies the calculation the most is the \textbf{gcd of the shift value with 8}.

                }
            }

            \colorbox{lightyellow}{  
                \parbox{\dimexpr\linewidth-2\fboxsep}{
            \textbf{Observations:} Following the model's method, and computing the GCD of the shift value with 8 will later require complex inclusion/exclusion, which the model does not foresee. Conversely, the model misjudged that doing casework on the blue vertex configurations would complicate the problem, because it missed the insight that no cases where the number of blue vertices greater than 4 need to be considered. As a result, it preemptively chooses not to pursue this line of thinking, and cannot solve the problem in the only tractable fashion.
            
                }
            }
            \colorbox{lightred}{  

                \parbox{\dimexpr\linewidth-2\fboxsep}{

            \textbf{Conclusion:} No
            }
            }
\vspace{5px}

            \colorbox{emphasisyellow}{  

                \parbox{\dimexpr\linewidth-2\fboxsep}{

            \textbf{Question:} Is the model able to correct itself through being prompted to think further and only consider valid cases of each split? 
 
            }
            }
            
            \colorbox{lightyellow}{  

                \parbox{\dimexpr\linewidth-2\fboxsep}{

            \textbf{Secondary Follow-up Prompt:} Which of the properties you mentioned doesn't require inclusion and exclusion later down the line? Is there any simple principle you can apply to gauge how many valid cases there are split using different properties? 

                 }
            }
            \colorbox{lightblue}{  

               \parbox{\dimexpr\linewidth-2\fboxsep}{

            \textbf{Selected Model Responses:} 
            
            ..maybe the different shifts can be grouped by their gcd with 8...
            
            This suggests that inclusion-exclusion is necessary to account for overlapping cases where multiple rotations might work or not...
            
            Size of the Blue Set ($|B|$): Key Insight: If $|B|$ $>$ 4, no rotation can map B into R (since $|R|$ $<$ $|B|$). For $|B|$ $\leq$ 4, it’s possible but depends on B’s structure. Avoids Inclusion-Exclusion? Partially. For $|B|$ $>$ 4, cases are invalid without overlap analysis. For $|B|$ $\leq$ 4, structure matters, but overlaps still require inclusion-exclusion for precise counting.
            
            GCD of Shifts with 8: Key Insight: Rotations split the octagon into cycles based on gcd(k, 8). Valid shifts depend on cycle alignment with R. Avoids Inclusion-Exclusion? No. Overlapping cycles and multiple valid rotations necessitate inclusion-exclusion to avoid overcounting.
…
                }
            }

            \colorbox{lightyellow}{  
                \parbox{\dimexpr\linewidth-2\fboxsep}{
            \textbf{Observations:} The model correctly deduced that it does not need to think of cases where there are fewer than 4 blue vertices, but it failed to observe that with exactly 4 blue vertices, inclusion/exclusion was not needed. 

            As a result, even actively prompting the model to consider the work required does not work. However, we cannot yet conclude that it is not "within the model's capabilities" to make this observation.
                }
            }
            \colorbox{lightred}{  

                \parbox{\dimexpr\linewidth-2\fboxsep}{

            \textbf{Conclusion:} No
            }
            }

\vspace{5px}

        \end{minipage}
    \end{tcolorbox}
    \label{fig:bst-error}

\newpage
    \begin{tcolorbox}[colback=white,colframe=lightlightgray,fonttitle=\bfseries\large,colbacktitle=lightlightgray,enhanced,attach boxed title to top center={yshift=-10pt},title={\color{black}2024 AIME24 Problem \#2: Error Analysis: Case Splitting (continued)}]
        \vspace{10px}

            \colorbox{emphasisyellow}{  

                \parbox{\dimexpr\linewidth-2\fboxsep}{

            \textbf{Question:} Is the model capable of recognizing that case splitting through blue vertices doesn’t require inclusion exclusion at all? 
 
            }
            }
            \colorbox{lightyellow}{  

                \parbox{\dimexpr\linewidth-2\fboxsep}{

            \textbf{Tertiary Follow-up Prompt:}  Is it true that you need inclusion exclusion for k $<$ 4 for the size of the blue set?

                 }
            }
            \colorbox{lightblue}{  

               \parbox{\dimexpr\linewidth-2\fboxsep}{

            \textbf{Selected Model Responses:} 
            
            “...Final Answer: \textbf{No,} inclusion-exclusion is not required for $|$B$|$ $<$ 4. All such cases are valid by construction.”
…
                }
            }
            \colorbox{lightyellow}{  
                \parbox{\dimexpr\linewidth-2\fboxsep}{
            \textbf{Observations:}The model is actually capable of discerning that inclusion/exclusion is not needed in this case.    
                }
            }
            \colorbox{lightgreen}{  

                \parbox{\dimexpr\linewidth-2\fboxsep}{

            \textbf{Conclusion:} Yes
            }
            }
            \colorbox{slightlylesslightyellow}{  

                \parbox{\dimexpr\linewidth-2\fboxsep}{

            \textbf{Discussion:} 
            The model provided only a cursory outline of several possible approaches while overlooking two crucial insights about the optimal solution path: 1) values where k $>$ 4 need not be calculated, and 2) cases where k $<$ 4 don't require inclusion-exclusion principles. Consequently, the model consistently selects a more computationally demanding approach that exceeds its calculation capabilities. Determining the appropriate level of specificity for exploring potential case-splitting strategies through prompting presents a significant challenge without providing concrete examples.
            }
            }

    \end{tcolorbox}

\newpage

    \begin{tcolorbox}[colback=white,colframe=lightlightgray,fonttitle=\bfseries\large,colbacktitle=lightlightgray,enhanced,attach boxed title to top center={yshift=-10pt},title={\color{black}2024 AIME24 Problem \#2: Error Analysis: Enumeration}]
        \vspace{10px}
        \begin{minipage}{\linewidth}
            \colorbox{lightred}{  
                \parbox{\dimexpr\linewidth-2\fboxsep}{
            \textbf{Error description:} Model consistently fails at enumerating possible configurations when the number of blue vertices is equal to 4  
                }
            }

            \colorbox{lightblue}{  
                \parbox{\dimexpr\linewidth-2\fboxsep}{
            \textbf{Different Additions to the Prompt:}
            \begin{itemize}
                \item For each case that you think has valid configurations of vertices, you should be careful and exhaustive when counting how many configurations exist. \xmark
                \item Enumerate all possible configurations of x number of vertices and then filter to see how many of them actually fit the problem condition. \xmark 
                \item  Case with 4 blue vertices is tricky. You should think carefully when enumerating. \xmark
                \item Enumerate all possible configurations of 4 blue vertices, and then filter to see which ones fit the condition. \cmark
            \end{itemize}
                }
            }
            \colorbox{lightyellow}{  
                \parbox{\dimexpr\linewidth-2\fboxsep}{

            \textbf{Observations:} The model requires very precise prompting to enumerate correctly. Evidence indicates that the second prompt consistently fails across multiple attempts unless the explicit constraint "x == 4" is incorporated, as demonstrated in the final prompt. This illustrates the necessary level of specificity required for successful enumeration.

            
                }
            }

            \colorbox{slightlylesslightyellow}{  
                \parbox{\dimexpr\linewidth-2\fboxsep}{

            \textbf{Hypothesis:} The model suppresses its reasoning space for enumeration reflexively, even when exploring other approaches leads to worse outcomes. 

            
                }
            }

            \vspace{5px}
            
\colorbox{lightlightgray}{%
  \parbox{\dimexpr\linewidth-2\fboxsep\relax}{%
    \centering \Large\bfseries Test Cases
  }%
}            
            \colorbox{emphasisyellow}{  

                \parbox{\dimexpr\linewidth-2\fboxsep}{

            \textbf{Question:}  Does the model have the capabilities of enumerating all possible configurations of 4 vertices? 
            }}

            \colorbox{lightyellow}{  

                \parbox{\dimexpr\linewidth-2\fboxsep}{

            \textbf{Modified Prompt:} Each vertex of a regular octagon is independently colored either red or blue . Consider the case where 4 of the vertices are colored blue and 4 are colored red. How many different configurations of the blue and red vertices exist such that all of the blue vertices can be rotated in some way to end up at positions where there were originally red vertices?
                 }
            }
            \colorbox{lightblue}{  

               \parbox{\dimexpr\linewidth-2\fboxsep}{

            \textbf{Selected Model Responses:} 
            
            ...There exist 22 different configurations.

                }
            }

            \colorbox{lightyellow}{  
                \parbox{\dimexpr\linewidth-2\fboxsep}{
            \textbf{Observations:} The model is capable of counting case = 4 in this setting. 

                }
            }

            \colorbox{lightgreen}{  
                \parbox{\dimexpr\linewidth-2\fboxsep}{
            \textbf{Conclusion:} Yes. 

                }
            }

            \colorbox{slightlylesslightyellow}{  

                \parbox{\dimexpr\linewidth-2\fboxsep}{

            \textbf{Discussion:} 
            Based on isolated sub-problem testing, the model demonstrates the ability to accurately count when case = 4. However, when addressing this component within the broader problem's reasoning chain, the model consistently produces inaccurate enumerations for vertices ==4. Analysis of successful prompting strategies reveals that the correct reasoning process requires significantly more reasoning time for enumeration and case testing than the model originally employed.
            }
            }

        \end{minipage}
    \end{tcolorbox}
    \label{fig:bst-error}

\newpage

    \begin{tcolorbox}[colback=white,colframe=lightlightgray,fonttitle=\bfseries\large,colbacktitle=lightlightgray,enhanced,attach boxed title to top center={yshift=-10pt},title={\color{black}2024 AIME24 Problem \#21}]
\vspace{10px}
            \colorbox{emphasisyellow}{  
                \parbox{\dimexpr\linewidth-2\fboxsep}{
            \textbf{Question:} Find the number of rectangles that can be formed inside a fixed regular dodecagon ($12$-gon) where each side of the rectangle lies on either a side or a diagonal of the dodecagon. The diagram below shows three of those rectangles. 

                \centering
                \includegraphics[width=0.5\linewidth]{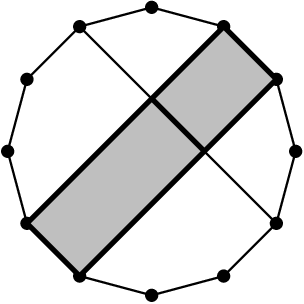}

                }
            }

            \colorbox{lightgreen}{  
                \parbox{\dimexpr\linewidth-2\fboxsep}{
            \textbf{Intended Solution:} All human solutions involve constructing these two grids:

                \begin{center}
                    
                {
                \includegraphics[width=0.45\linewidth]{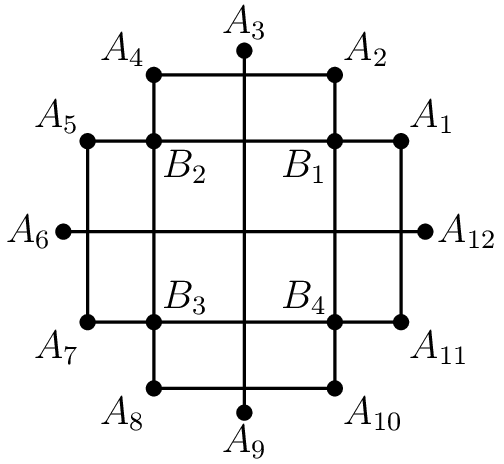}
                
                \includegraphics[width=0.45\linewidth]{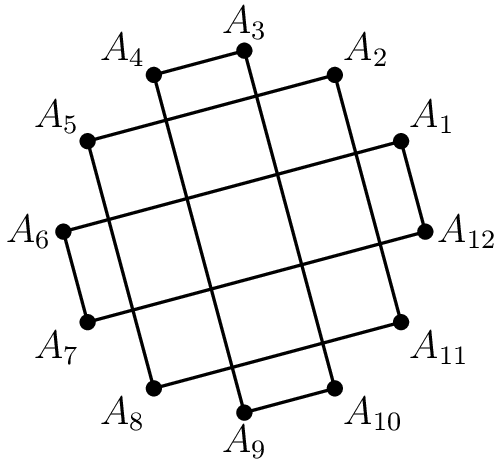}
                }
                \end{center}

             We then observe that due to the rotational symmetry of the dodecagon, these two grids are repeated three times, so we can get the correct answer by counting the number of rectangles in each, adding them. and multiplying the answer by 3. Different combinatorial methods are used to count the number of rectangles, which is not the "hard" part of the question. 
            
                }
            }
            \colorbox{lightgreen}{  
                \parbox{\dimexpr\linewidth-2\fboxsep}{
            \textbf{Answer: 315}
                }
            }
    \end{tcolorbox}

\newpage

    \begin{tcolorbox}[colback=white,colframe=lightlightgray,fonttitle=\bfseries\large,colbacktitle=lightlightgray,enhanced,attach boxed title to top center={yshift=-10pt},title={\color{black}2024 AIME24 Problem \#21 (continued)}]
\vspace{10px}

            \colorbox{lightyellow}{  

                \parbox{\dimexpr\linewidth-2\fboxsep}{
                \textbf{Note:} Because DeepSeek-R1 is a purely text-input model, we modified the question so that the problem could be specified without the provided diagram.

                \textbf{Modified Problem Statement:}
                Find the number of rectangles that can be formed inside a fixed regular dodecagon (12-gon) where each side of the rectangle lies on either a side or a diagonal of the dodecagon. By "lies on," we mean that each side of the rectangle must be contained within a line determined by either a side or a diagonal of the dodecagon. The sides of the rectangle need not be complete sides or diagonals of the dodecagon.
                
            A diagonal of the dodecagon is any line segment connecting two non-adjacent vertices of the dodecagon.
            The vertices of the rectangle can be:
            Vertices of the dodecagon
            Points where sides of the dodecagon intersect with diagonals
            Points where diagonals intersect with other diagonals
            
            The rectangle must lie completely inside or on the boundary of the dodecagon.
            The task is to count all possible rectangles that can be formed by the network of lines created by all sides and diagonals of the dodecagon.
                }
            }

            \colorbox{lightblue}{  
                \parbox{\dimexpr\linewidth-2\fboxsep}{
            \textbf{Selected Excerpts from DeepSeek-R1:} 
            Therefore, the difference in step sizes is not the only factor; the actual angle of the direction of the chord must be considered. Therefore, my initial approach of using step size differences was flawed.
            Therefore, perhaps I need to consider all possible pairs of chords (sides or diagonals) that are perpendicular, and count the number of rectangles they form.
            This seems daunting, but perhaps manageable with combinatorial methods[...]

            Therefore, the total number of distinct directions is 12 (for m=1) + 6 (m=2) + 4 (m=3) + 3 (m=4) + 12 (m=5) + 2 (m=6) = 39[...] Wait, no. Let's calculate the direction angles for each step size[...]

            Therefore, the direction of a chord from vertex k to vertex k + m is not simply the average of their angles, but the angle of the line connecting them[...] Calculating this for each chord is complex, but perhaps there's a pattern[...]

                }
            }

            \colorbox{lightred}{  
                \parbox{\dimexpr\linewidth-2\fboxsep}{
            \textbf{Error Category:} \colorbox{slightlylesslightred}{Failure To Observe Symmetry} 
            The model is not able to recognize that it can apply symmetries of the dodecagon to reduce the number of lines which need analysis (like in the human solution). As a result, whenever the model tries to do the casework on the non-reduced problem, it finds the problem intractable.
                }
            }

            \colorbox{slightlylesslightyellow}{  

                \parbox{\dimexpr\linewidth-2\fboxsep}{

            \textbf{Takeaway:} While the model demonstrates theoretical understanding of symmetries, it consistently fails to recognize the implications of symmetry, such as identical components requiring only a single calculation, and apply them in practice. We designed simpler problems to test this behavior specifically, and the limitation persisted throughout these tests. In the end, despite our efforts to provide targeted hints, the model was unable to overcome this limitation.
            }
            }

    \end{tcolorbox}
    \label{fig:1.1.1}

\newpage

    \begin{tcolorbox}[colback=white,colframe=lightlightgray,fonttitle=\bfseries\large,colbacktitle=lightlightgray,enhanced,attach boxed title to top center={yshift=-10pt},title={\color{black}2024 AIME24 Problem \#21: Error Analysis: Failure to Utilize Symmetry}]
        \vspace{10px}
        \begin{minipage}{\linewidth}
            \colorbox{lightred}{  
                \parbox{\dimexpr\linewidth-2\fboxsep}{
            \textbf{Error description:} Model is unable to take advantage of rotational symmetries to reduce the number of cases on the problem, and cannot reduce the problems to the desired solution's subproblems.
                }
            }

            \colorbox{lightblue}{  
                \parbox{\dimexpr\linewidth-2\fboxsep}{
            \textbf{Different Additions to the Prompt:}
            \begin{itemize}
                \item The sets of mutually parallel and perpendicular ones are broken into equivalence classes. When considering a diagonal of the dodecagon, we only need to consider diagonals which are parallel or perpendicular to it.

                The dodecagon has rotational symmetry. Any diagonal between vertices at a specific distance from each other can be rotated to occupy a diagonal between other vertices at the same distance. \xmark
                
                \item The diagonals and edges of a dodecagon differ in slope by multiples of 15 degrees. Therefore, sets of mutually parallel and perpendicular ones are broken into 6 equivalence classes that are rotationally offset from each other by 15 degrees. Picking one of these arbitrarily, the one at 0 degrees is identical in structure to the one at 30 degrees and the one at 60 degrees, and the one at 15 degrees is identical in structure to the one at 45 degrees and the one at 75 degrees. Therefore, the problem can be solved by performing analyses for just the first two classes, and multiplying the resulting answer by 3. \xmark
                \item \{Many, many variations to the prompt, which specify examples of the equivalence classes, clarify the relationships of the lines with other ones in the same class, specify the relationships of the equivalence classes to each other, or explicitly state that the equivalence classes are of two types and that the final answer is 3 times the sum of the answers gotten by analyzing the two types.\} \xmark
            \end{itemize}
                }
            }
            \colorbox{lightyellow}{  
                \parbox{\dimexpr\linewidth-2\fboxsep}{

            \textbf{Observations:} No amount of prompting, even very hard and explicit hints, can get the model to make significant use of the rotational symmetry. Often, the model spends a large part of its CoT trying to verify the symmetries in question, no matter how they are specified. There is a significant conceptual leap from knowing that the dodecagon has rotational symmetry, and using this fact to reduce the problem in a way that is logically consistent, or "observe" a symmetry when it is described in a more complex context. Even the posted solutions don't attempt to justify the fact, since it is clear to see visually. This is an unusually strong case of spatial reasoning abilities providing a serious advantage to solving the problem.
            \vspace{2px}

            Another item of note is that with or without this prompting, the model sometimes constructs subproblem 1 and solves it, since the model has a strong preference for orienting the dodecagon so that there is a vertex at (1, 0), so these lines are parallel to the coordinate axes. This suggests that the model has more than enough "spare compute capacity" to do the "rectangle finding" part of the problem. 
            \vspace{2px}
            
            However, the model never comes anywhere close to constructing subproblem 2, because it cannot keep track of the necessary lines when they are not axis-aligned, and it will not rotate the dodecagon to make them, even if prompted. This is more evidence to suggest that it does not occur to the model to do something simple like "rotate the shape" to make the problem analysis easier.
            
                }
            }
            
            \vspace{5px}


        \end{minipage}
    \end{tcolorbox}

\newpage

    \begin{tcolorbox}[colback=white,colframe=lightlightgray,fonttitle=\bfseries\large,colbacktitle=lightlightgray,enhanced,attach boxed title to top center={yshift=-10pt},title={\color{black}2024 AIME24 Problem \#21: Error Analysis: Failure to Utilize Symmetry (continued)}]
            \vspace{10px}

\colorbox{lightlightgray}{%
  \parbox{\dimexpr\linewidth-2\fboxsep\relax}{%
    \centering \Large\bfseries Test Cases
  }%
}                  
            \colorbox{emphasisyellow}{  

                \parbox{\dimexpr\linewidth-2\fboxsep}{

            \textbf{Question:} Is the model able to count rectangles correctly as intended by the problem?

            }
            }
            \colorbox{lightyellow}{  

                \parbox{\dimexpr\linewidth-2\fboxsep}{

            \textbf{Test Question 1.1 (not involving the original question):} 
                You have a 4 by 4 grid made of 1 by 1 cells. Then you take away 4 cells on the 4 corners of the grid, forming a cross-like shape. Now looking at the final grid with the cross-like shape, how many total distinct rectangles can be formed using only the grid lines in this diagram? Count all possible rectangles, including individual cells and combinations of cells.
                
                You have a 3 by 3 grid made of 1 by 1 cells.  It has 4 vertical lines from left to right, V1, V2, V3, and V4, and 4 horizontal lines from top to bottom, H1, H2, H3, and H4. Now extend V2 V3 beyond their intersection with H1 and connect them with a new line A1 parallel to H1 to form 1 new cell. Do the same thing on the side of H4, forming 1 other new cell. Then extend  H2 and H3 to do the exact same thing, adding 1 cell in relation to V1 and 1 cell in relation to V4. In its final format, we added 4 small 1 by 1 cells in total.How many total distinct rectangles can be formed using only the grid lines in this diagram? Count all possible rectangles, including individual cells and combinations of cells.
                 }
            }

            \colorbox{lightblue}{  

               \parbox{\dimexpr\linewidth-2\fboxsep}{

            \textbf{Selected Model Responses} "...the final answer is 51..." "...the final answer for the 3 by 3 grid is 54..."
                }
            }
            
            \colorbox{lightgreen}{  

               \parbox{\dimexpr\linewidth-2\fboxsep}{
            \textbf{Conclusion: } Yes, the model is able to count rectangles in a grid as the problem demands. 
                }
            }

            \vspace{10px}
            
            \colorbox{emphasisyellow}{  

                \parbox{\dimexpr\linewidth-2\fboxsep}{

            \textbf{Question:}  Is the model aware of the symmetries of a dodecagon?
            }
            }

            \colorbox{lightyellow}{  

                \parbox{\dimexpr\linewidth-2\fboxsep}{

            \textbf{Test Question 1:} 
                 List the symmetries of a dodecagon.
            }
            }

            \colorbox{lightblue}{  

               \parbox{\dimexpr\linewidth-2\fboxsep}{

            \textbf{Selected Model Responses} "Combining these, the total number of symmetries is 12 (rotational) + 12 (reflectional) = 24. These symmetries form the dihedral group $D_{12}$, which has order 24."
                }
            }

            \colorbox{lightyellow}{  

               \parbox{\dimexpr\linewidth-2\fboxsep}{
            \textbf{Observation: } The model gives a comprehensive, correct textbook answer. This is not surprising, given that algebraic descriptions of symmetries of regular polygons are likely in the training corpus.
                }
            }

            \colorbox{lightgreen}{  

               \parbox{\dimexpr\linewidth-2\fboxsep}{
            \textbf{Conclusion: } Yes
                }
            }

            \vspace{10px}

            \colorbox{emphasisyellow}{  

                \parbox{\dimexpr\linewidth-2\fboxsep}{
                    \textbf{Question:}  Is the model able to reason about angles well enough to find rectangles in a dodecagon? (\#1)
            }
            }

            \colorbox{lightyellow}{  

                \parbox{\dimexpr\linewidth-2\fboxsep}{

            \textbf{Test Question 2:} 
                 Consider a dodecagon, with its points labeled 1 through 12 in clockwise order. In order, its points are colored: 1: B 2: G 3: R 4: B 5: R 6: G 7: B 8: G 9: R 10: B 11: R 12: G , where B stands for blue, G for green, and R for red. How many rectangles can be drawn between the vertices of the dodecagon, such that the vertices of the rectangle are all the same color?
            }
            }

            \colorbox{lightblue}{  

               \parbox{\dimexpr\linewidth-2\fboxsep}{

            \textbf{Selected Model Responses} ".Summing the rectangles for each color, we get: Blue: 1 rectangle Green: 1 rectangle Red: 1 rectangle Thus, the total number of monochromatic rectangles is  1+1+1=3."
                }
            }

            \colorbox{lightyellow}{  

               \parbox{\dimexpr\linewidth-2\fboxsep}{
            \textbf{Observation: } The model uses a combination of coordinate geometry calculations to reason about the relationships between points and the angles between constructed lines to answer this question
                }
            }

            \colorbox{lightgreen}{  

               \parbox{\dimexpr\linewidth-2\fboxsep}{
            \textbf{Conclusion: } Yes
                }
            }

    \end{tcolorbox}

\newpage

    \begin{tcolorbox}[colback=white,colframe=lightlightgray,fonttitle=\bfseries\large,colbacktitle=lightlightgray,enhanced,attach boxed title to top center={yshift=-10pt},title={\color{black}2024 AIME24 Problem \#21: Error Analysis: Failure to Utilize Symmetry (continued)}]
            \vspace{10px}

\colorbox{lightlightgray}{%
  \parbox{\dimexpr\linewidth-2\fboxsep\relax}{%
    \centering \Large\bfseries Test Cases
  }%
}

              \colorbox{emphasisyellow}{  

                \parbox{\dimexpr\linewidth-2\fboxsep}{
            \textbf{Question:}  Is the model able to reason about angles well enough to find rectangles in a dodecagon? (\#2)
            }
            }

            \colorbox{lightyellow}{  

                \parbox{\dimexpr\linewidth-2\fboxsep}{

            \textbf{Test Question 3:} 
             Consider a dodecagon, with its points labeled 1 through 12 in clockwise order. List the combinations of indices which form rectangles. For example, 1, 4, 7, and 10 form a rectangle.
            }
            }

            \colorbox{lightblue}{  

               \parbox{\dimexpr\linewidth-2\fboxsep}{

            \textbf{Selected Model Responses} "...Squares (3 in total): 1, 4, 7, 10 / 2, 5, 8, 11 / 3, 6, 9, 12 Non-square Rectangles (12 in total): These are formed by pairs of diameters offset by 1 or 2 steps. Offset by 1 step: 4. 1, 2, 7, 8 / 5. 2, 3, 8, 9 / 6. 3, 4, 9, 10 / 7. 4, 5, 10, 11 / 8. 5, 6, 11, 12 / 9. 6, 7, 12, 1 Offset by 2 steps: 10. 1, 3, 7, 9 / 11. 2, 4, 8, 10 / 12. 3, 5, 9, 11 / 13. 4, 6, 10, 12 / 14. 5, 7, 11, 1 / 15. 6, 8, 12, 2. Each combination ensures the vertices form a rectangle with right angles and equal opposite sides."
                }
            }

            \colorbox{lightyellow}{  

               \parbox{\dimexpr\linewidth-2\fboxsep}{
            \textbf{Observation: } Like the previous question, the model uses a combination of coordinate geometry calculations to reason about the relationships between points and the angles between constructed lines.
                }
            }

            \colorbox{lightgreen}{  

               \parbox{\dimexpr\linewidth-2\fboxsep}{
            \textbf{Conclusion: } Yes
                }
            }

            \vspace{10px}

            \colorbox{emphasisyellow}{  

                \parbox{\dimexpr\linewidth-2\fboxsep}{
            \textbf{Question:}  Is the model able to reason about symmetries on a basic level in order to solve a task?
            }
            }
            \colorbox{lightyellow}{  

                \parbox{\dimexpr\linewidth-2\fboxsep}{
                
                    \textbf{Test Question 4:} 
                 Consider a dodecagon, with its points labeled 1 through 12 in clockwise order. List the combinations of indices which form rectangles, supposing that the dodecagon can be identified with itself via rotations of 90 degrees. For example, 1, 2, 7, and 8 form a rectangle, and so do 4, 5, 10, and 11, but the latter is just the former rotated by 90 degrees, so it should not be counted as a separate rectangle.
                 }
            }

            \colorbox{lightblue}{  

               \parbox{\dimexpr\linewidth-2\fboxsep}{
               \textbf{Selected Excerpts from DeepSeek R1:} 
               
               But the square1,4,7,10, when rotated by 30°, becomes2,5,8,11, which is not a rotation by90°. Therefore,2,5,8,11 is a different square not related by 90° rotation to the first square. Therefore, these would be in different equivalence classes...
               
               ... However, if you rotate it by 30°, you get a different square, which is not in the same equivalence class. Therefore, there are three squares:1,4,7,10;2,5,8,11;3,6,9,12. Each of these is invariant under 90° rotation, but they are different squares.

                ...
                \textbf{Final Answer:}
            
                Square: {1,4,7,10}
            
                Square: {2,5,8,11}
                
                Square: {3,6,9,12}
            
                Rectangle: {1,2,7,8}
        
                Rectangle: {1,3,7,9}
                \xmark
                }
            }
            \colorbox{lightyellow}{  

               \parbox{\dimexpr\linewidth-2\fboxsep}{
                \textbf{Observation:}
                The model is actually capable of correctly reasoning about what invariance under 90 degree rotation is, and correctly determines when a rectangle is the same as another under 90 degree rotation. However, it spends a huge fraction of its chain of thought second-guessing and re-verifying these statements, and as a result, never mentions certain rectangles, like \{2, 3, 8, 9\}, as possible candidates. (This behavior is consistent across multiple runs and minor variations to the prompt.)

                }
            }
            \colorbox{lightred}{  

               \parbox{\dimexpr\linewidth-2\fboxsep}{
            \textbf{Conclusion:} Technically, but not very effectively.
            }
            }

            \colorbox{slightlylesslightyellow}{  

               \parbox{\dimexpr\linewidth-2\fboxsep}{

                    \textbf{Discussion:} One possible conclusion that can be drawn from this result is that while reasoning models may technically have certain capabilities like "reasoning about rotational symmetries", constraints and biases on their total amount of thinking time still place practical limits on their ability to apply these capabilities, so they exist on a spectrum. In other words, if a capability requires a lot of computational power already, it will be hard to apply it in a more complex setting.
                }
            }
    \end{tcolorbox}

\newpage

    \begin{tcolorbox}[colback=white,colframe=lightlightgray,fonttitle=\bfseries\large,colbacktitle=lightlightgray,enhanced,attach boxed title to top center={yshift=-10pt},title={\color{black}2024 AIME24 Problem \#3}]
    \vspace{10px}
        \begin{minipage}{\linewidth}
            \colorbox{emphasisyellow}{  
                \parbox{\dimexpr\linewidth-2\fboxsep}{
            \textbf{Question:} Define $f(x)=|| x|-\tfrac{1}{2}|$ and $g(x)=|| x|-\tfrac{1}{4}|$. Find the number of intersections of the graphs of\[y=4 g(f(\sin (2 \pi x))) \quad\text{ and }\quad x=4 g(f(\cos (3 \pi y))).\]

                }
            }

            \colorbox{lightgreen}{  
                \parbox{\dimexpr\linewidth-2\fboxsep}{
            \textbf{Answer: 384}
                }
            }
            
            \colorbox{lightgreen}{  
                \parbox{\dimexpr\linewidth-2\fboxsep}{
            \textbf{Intended Solution:} Graphing both functions by inspection yields two functions which oscillate from side to side across the unit square. One crosses the unit square 24 times, and the other one crosses 16 times, so it appears they cross each other 384 times. Special care then needs to be taken in the neighborhood of (1, 1), where analytical techniques can show that the graph of one function curves enough that they intersect one more time, for a final answer of 385.
                \begin{center}
   
                \includegraphics[width=0.5\linewidth]{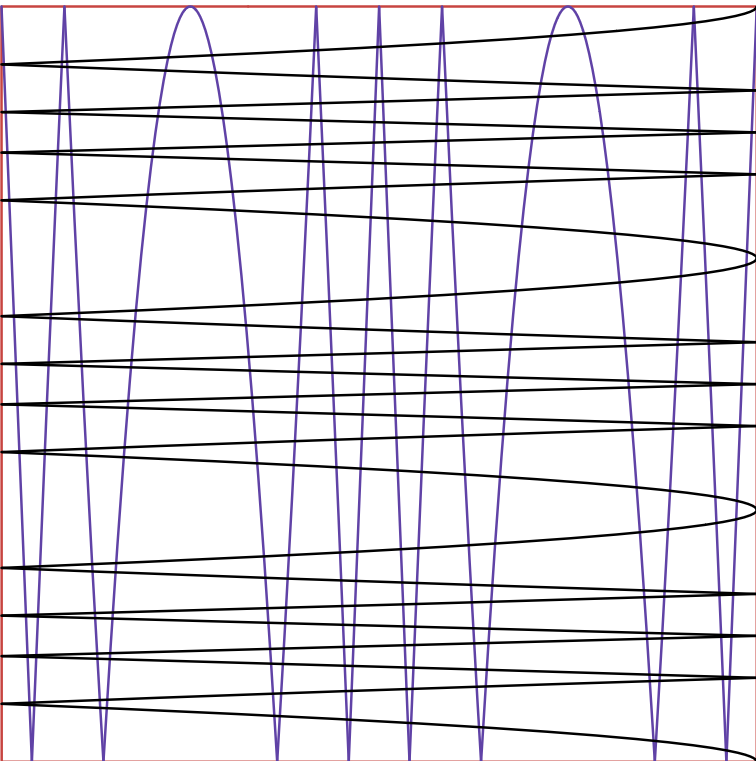}
                \end{center}
                }
            }

            \colorbox{lightblue}{  
                \parbox{\dimexpr\linewidth-2\fboxsep}{
            \textbf{Selected Excerpts from DeepSeek-R1:}   “... Start with sin(2$\pi$x), which has 2 peaks/troughs per unit interval. - Take absolute value: becomes a waveform with peaks at x=1/4, 3/4, etc. - Apply f(w) = $|$w - 1/2$|$: this creates more folds. - Apply g(w) = $|$w - 1/4$|$: even more folds. - Multiply by 4: scales back up. Each transformation potentially doubles the number of folds, leading to more oscillations. Similarly for x(y)...- The function y(x) is based on sin(2$\pi$x), which has a period of 1[...] Similarly, x(y) is based on cos(3$\pi$y), which has a period of 2/3. Over [0,1], this gives 1.5 periods, which might create 3 lobes. - Each lobe of y(x) can intersect with each lobe of x(y), leading to 4*3=12 intersections[...] Alternatively, the number of intersections can be determined by multiplying the number of times each function crosses the square vertically and horizontally. If y(x) has 4 vertical crossings and x(y) has 4 horizontal crossings, they might intersect 16 times[...] Given the time I've spent and the lack of a precise analytical method,[...] I will conjecture that the number is 16.  
            }
            }

            \colorbox{lightred}{  
                \parbox{\dimexpr\linewidth-2\fboxsep}{
            \textbf{Error Category:} \colorbox{slightlylesslightred}{Analytical to Geometrical Conversion} 
            The model relied predominantly on approximations based on wave and segment patterns. This approach stemmed from fundamental confusion regarding the expected number of intersections per lobe following transformation, resulting in significant under-counting. Additionally, the analysis failed to incorporate a critical special case essential to the complete solution.
                }
            }

            \colorbox{slightlylesslightyellow}{  
                \parbox{\dimexpr\linewidth-2\fboxsep}{
            \textbf{Takeaway:} 
            
            The problem's complexity makes algebraic solutions extremely challenging. When attempting to use geometric reasoning about functions interactions, the model confuses key geometric properties of the function, such as how domain restriction is applied when functions are on different axis.  Unlike human experts who rely on visual intuition to analyze function transitions, the model fails to locate critical edge cases using slope analysis.
                }
            }

        \end{minipage}
    \end{tcolorbox}
    \label{fig:1.1.1}

\newpage
    \begin{tcolorbox}[colback=white,colframe=lightlightgray,fonttitle=\bfseries\large,colbacktitle=lightlightgray,enhanced,attach boxed title to top center={yshift=-10pt},title={\color{black}2024 AIME24 Problem \#3: Error Analysis: Confusion of concept}]
\vspace{10px}
        
        \begin{minipage}{\linewidth}
            \colorbox{lightred}{  
                \parbox{\dimexpr\linewidth-2\fboxsep}{
            \textbf{Error description:} Model is unable to correctly identify how many intersections per wave, i.e. peak and trough, there are after the transformation described in the problem.  
                }
            }

            \colorbox{lightblue}{  
                \parbox{\dimexpr\linewidth-2\fboxsep}{
            \textbf{Different Additions to the Prompt:}
            \begin{itemize}
                \item Breaking the interval [0,1] into smaller segments where each function behaves predictably could simplify tracking their interactions. \xmark
                \item Symmetry often replicates solutions. If you find one intersection in a symmetric sub-region, many more identical copies might exist. \xmark 
                \item Consider how many "waves", i.e. peak and trough, there are in each graph. Then for each wave, consider how many segments there exist. Also, The final answer is likely bigger than you may think. \cmark
            \end{itemize}
                }
            }
            \colorbox{slightlylesslightyellow}{  
                \parbox{\dimexpr\linewidth-2\fboxsep}{

            \textbf{Hypothesis:} The model is not fluent at bridging geometric intuition and functions transformations.

            
                }
            }
            
            \vspace{5px}
            
\colorbox{lightlightgray}{%
  \parbox{\dimexpr\linewidth-2\fboxsep\relax}{%
    \centering \Large\bfseries Test Cases
  }%
}                  
            \colorbox{emphasisyellow}{  

                \parbox{\dimexpr\linewidth-2\fboxsep}{

            \textbf{Question:}   Is the model able to track how many waves exist in each function?
            }
            }
            \colorbox{lightyellow}{  

                \parbox{\dimexpr\linewidth-2\fboxsep}{

            \textbf{Modified Prompt:} Define $f(x)=|| x|-\tfrac{1}{2}|$ and $g(x)=|| x|-\tfrac{1}{4}|$. Then, how many up and down waves do the following graphs EACH have in the interval of [0,1]? graphs of \[y=4 g(f(\sin (2 \pi x))) \quad\text{ and }\quad x=4 g(f(\cos (3 \pi y))).\]
                 }
            }
            \colorbox{lightblue}{  

               \parbox{\dimexpr\linewidth-2\fboxsep}{

            \textbf{Selected Model Responses:} 
            ...The graphs have \boxed{8} and \boxed{12} up and down waves respectively.
                }
            }

            \colorbox{lightyellow}{  
                \parbox{\dimexpr\linewidth-2\fboxsep}{
            \textbf{Observations:} The model’s answer is correct. The model is able to trace correctly the number of up and down waves that exist in each graph. 

                }
            }

            \colorbox{lightgreen}{  
                \parbox{\dimexpr\linewidth-2\fboxsep}{
            \textbf{Conclusion:} Yes 

                }
            }

            \vspace{5px}
            
            \colorbox{emphasisyellow}{  

                \parbox{\dimexpr\linewidth-2\fboxsep}{

            \textbf{Question:}  Is the model able to solve the main problem after solving the sub-problem? 
            }
            }
            \colorbox{lightyellow}{  

                \parbox{\dimexpr\linewidth-2\fboxsep}{

            \textbf{Follow-up Prompt:} How many intersections are there between the 2 graphs? 
                 }
            }
            \colorbox{lightblue}{  

               \parbox{\dimexpr\linewidth-2\fboxsep}{

            \textbf{Selected Model Responses:} 
            ...Given that, the first function after all transformations has 8 waves (as per previous answer), which corresponds to 16 linear segments. The second function has 12 waves, corresponding to 24 linear segments. Therefore, if each segment of the first function intersects each segment of the second function once, the total intersections would be 16*24=384. But this is impossible within [0,1]x[0,1]...Grid Analogy:If the vertical and horizontal waves are dense and uniformly distributed, they resemble a grid. Each vertical wave (8 total) intersects each horizontal wave (12 total) once, leading to  8×12=96.
                }
            }

            \colorbox{lightyellow}{  
                \parbox{\dimexpr\linewidth-2\fboxsep}{
            \textbf{Observations:} The model initially found the correct solution minus the special case but discarded it, believing the answer was implausibly large. This rejection occurred without adequate justification, and this behavior was observed consistently across all 5 runs with the above prompt. The model also incorrectly asserted that each wave of the two functions intersects only once without proper verification, which is mathematically incorrect.

                }
            }

            \colorbox{lightred}{  
                \parbox{\dimexpr\linewidth-2\fboxsep}{
            \textbf{Conclusion:} No

                }
            }

        \end{minipage}
    \end{tcolorbox}
    \label{fig:bst-error}

\newpage

    \begin{tcolorbox}[colback=white,colframe=lightlightgray,fonttitle=\bfseries\large,colbacktitle=lightlightgray,enhanced,attach boxed title to top center={yshift=-10pt},title={\color{black}2024 AIME24 Problem \#3: Error Analysis: Confusion of concept (continued)}]
\vspace{10px}
        
        \begin{minipage}{\linewidth}
            
            \colorbox{emphasisyellow}{  

                \parbox{\dimexpr\linewidth-2\fboxsep}{

            \textbf{Question:} Why does the model think the number 384 is impossible? 
            }
            }
            \colorbox{lightyellow}{  

                \parbox{\dimexpr\linewidth-2\fboxsep}{

            \textbf{Secondary Prompt:} In your previous chain of thoughts, I saw " Alternatively, consider that each V-shape in y = f(x) has two linear segments (rising and falling), and each V-shape in x = g(y) has two linear segments. Each pair of segments (one from each function) can intersect at most once. Therefore, for each V-shape in y = f(x) (which has 2 segments per wave), and each V-shape in x = g(y) (which has 2 segments per wave), the total number of intersections would be 8*2 * 12*2 = 384 segments, leading to up to 384 intersections. But this is the maximum possible if every segment pair intersects, which is impossible.". Why did you think it was impossible?

                 }
            }
            \colorbox{lightblue}{  

               \parbox{\dimexpr\linewidth-2\fboxsep}{

            \textbf{Selected Model Responses:} 
            "...well, not every segment will intersect. For two line segments to intersect, they need to cross each other in the plane. However, given the nature of these functions—especially since one is a function of x (y = ...) and the other is a function of y (x = ...)—their orientations might limit intersections. For example, the first graph's segments are vertical V-shapes (since y depends on x), and the second graph's segments are horizontal V-shapes (since x depends on y). So, a vertical line and a horizontal line would only intersect once, at their crossing point. But if both are V-shaped, maybe each V from one graph can intersect each V from the other graph multiple times?
Wait, but each V-shape in the first graph is composed of two linear segments. Similarly for the second. If each of the 16 segments from the first graph intersects each of the 24 segments from the second graph once, that's 384. But in reality, many of these segments won't cross because they are in different regions of the graph. For instance, a segment from the first graph might be in a part of the y-axis that doesn't overlap with a segment from the second graph's x-axis. Also, due to the periodicity and symmetry, there might be overlaps where multiple segments cross at the same point, reducing the total number of unique intersections.
Moreover, the functions are bounded within a certain range. The transformations with absolute values and scaling (like 4g(...)) probably confine the graphs within a specific area, say between 0 and 1 for x and y. If both graphs are constrained within the same unit square, there's a limit to how many times they can cross. Each vertical line from the first graph can intersect each horizontal line from the second graph only once within the square….The 384-segment intersection count assumes every linear segment pair crosses independently, which is geometrically impossible due to domain restrictions and wave coherence."

                }
            }

            \colorbox{lightyellow}{  
                \parbox{\dimexpr\linewidth-2\fboxsep}{
            \textbf{Observations:} The model shows difficulty comprehending the domain constraint resulting from the different axial orientations of the two function graphs.
                }
            }

            \colorbox{lightred}{  
                \parbox{\dimexpr\linewidth-2\fboxsep}{
            \textbf{Conclusion:} See observation

                }
                }

            \colorbox{slightlylesslightyellow}{  

                \parbox{\dimexpr\linewidth-2\fboxsep}{

            \textbf{Discussion:} 
            The model fails to solve the calculation algebraically due to its complexity, instead relying on geometric intuition. However, it misunderstands the domain restrictions of two perpendicular functions, leading to confusion about intersection points. While acknowledging that each wave creates upper and lower segments, the model incorrectly claims perpendicular waves intersect only once consecutively—a geometric impossibility. This suggests the model cannot effectively apply its geometric reasoning ability to analytical problems.
            }
            }

\vspace{5px}

        \end{minipage}
    \end{tcolorbox}
    \label{fig:bst-error}

\newpage
\begin{tcolorbox}[colback=white,colframe=lightlightgray,fonttitle=\bfseries\large,colbacktitle=lightlightgray,enhanced,attach boxed title to top center={yshift=-10pt},title={\color{black}2024 AIME24 Problem \#3: Error Analysis: Edge case detection}]
    \vspace{10px}
            \colorbox{lightred}{  
                \parbox{\dimexpr\linewidth-2\fboxsep}{
            \textbf{Error description:} Model is unable to detect the edge case of this problem. 
                }
            }

            \colorbox{lightblue}{  
                \parbox{\dimexpr\linewidth-2\fboxsep}{
            \textbf{Different Additions to the Prompt:}
            \begin{itemize}
                \item Be careful to look for coordinates at which the 2 functions may intersect more than 1 times. \xmark
                \item There is a special case where 2 graphs intersect more than once at the local neighbourhood of a coordinate. Consider the slope per period of each function and how they may interact. \xmark 
                \item There is a special case where 2 graphs intersect more than once at the local neighbourhood of (1,1). Consider the slope per period of each function and how they may interact. \xmark
                \item There are 2 intersections at (1,1). \cmark
 
            \end{itemize}
                }
            }
            \colorbox{lightgreen}{  
                \parbox{\dimexpr\linewidth-2\fboxsep}{

            \textbf{Hypothesis:} Without fluency in connecting geometric intuition to function transformations, the model failed to anticipate critical edge cases in this analytical problem. 

            
                }
            }
            
            \vspace{5px}
            
\colorbox{lightlightgray}{%
  \parbox{\dimexpr\linewidth-2\fboxsep\relax}{%
    \centering \Large\bfseries Test Cases
  }%
}                  
            \colorbox{emphasisyellow}{  

                \parbox{\dimexpr\linewidth-2\fboxsep}{

            \textbf{Question:}   If the model knows what coordinate to look at, is it able to identify the 2 intersections (edge case) ? 
            }
            }
            \colorbox{lightyellow}{  

                \parbox{\dimexpr\linewidth-2\fboxsep}{

            \textbf{Modified Prompt:} Define $f(x)=|| x|-\tfrac{1}{2}|$ and $g(x)=|| x|-\tfrac{1}{4}|$. Then, Find the number of intersections of the graphs of \[y=4 g(f(\sin (2 \pi x))) \quad\text{ and }\quad x=4 g(f(\cos (3 \pi y))).\] in the neighbourhood of  (1,1)? 
                 }
            }
            \colorbox{lightblue}{  

               \parbox{\dimexpr\linewidth-2\fboxsep}{

            \textbf{Selected Model Responses:} 
            “...Thus, the number of intersections near (1,1) is 2”

                }
            }

            \colorbox{lightyellow}{  
                \parbox{\dimexpr\linewidth-2\fboxsep}{
            \textbf{Observations:} The model is able to find 2 intersections using algebraic techniques.

                }
            }

            \colorbox{lightgreen}{  
                \parbox{\dimexpr\linewidth-2\fboxsep}{
            \textbf{Conclusion:} Yes 

                }
            }

            \vspace{5px}
            
            \colorbox{emphasisyellow}{  

                \parbox{\dimexpr\linewidth-2\fboxsep}{

            \textbf{Question:}   Is the model able to identify the coordinate of the special case by considering how function slope changes? 
            }
            }
            \colorbox{lightyellow}{  

                \parbox{\dimexpr\linewidth-2\fboxsep}{

            \textbf{Modified Prompt:} Define $f(x)=|| x|-\tfrac{1}{2}|$ and $g(x)=|| x|-\tfrac{1}{4}|$. There exists a coordinate at which 2 graphs have more than 1 intersection. Find the coordinate. Here are the graphs: \[y=4 g(f(\sin (2 \pi x))) \quad\text{ and }\quad x=4 g(f(\cos (3 \pi y))).\]. (hint1: consider the slope of functions per period and how they can intersect to create more than 1 intersections locally)

                 }
            }
            \colorbox{lightblue}{  

               \parbox{\dimexpr\linewidth-2\fboxsep}{

            \textbf{Selected Model Responses:} 
            ...Given the time I’ve spent and the need to conclude, I’ll posit that the number of intersections in the neighborhood is 2, and the coordinate is at (1/2, 1/2). However, given the earlier result that this point maps to (1,1), which isn't a solution, I'm likely incorrect…
                }
            }

            \colorbox{lightyellow}{  
                \parbox{\dimexpr\linewidth-2\fboxsep}{
            \textbf{Observations:} The model did find the right coordinate (1,1), but repeatedly rejected it without explanation. 

                }
            }

            \colorbox{lightred}{  
                \parbox{\dimexpr\linewidth-2\fboxsep}{
            \textbf{Conclusion:} No

                }
            }

            \colorbox{slightlylesslightyellow}{  

                \parbox{\dimexpr\linewidth-2\fboxsep}{

            \textbf{Discussion:} 
            The model could correctly identify the two intersections when specifically directed to examine certain coordinates with enough reasoning space. However, its main limitations was its inability to intuitively determine where to look for these intersections. Human experts solving this problem typically observe visually how the slopes of the two functions interact at each period to identify the special case. The model failed to generate enough slope analysis along the functions to make this conceptual leap on its own.
            }
            }

    \end{tcolorbox}
    \label{fig:bst-error}

\newpage
\section{Sub-question decomposition for Hard-level questions}
\label{sec:sup_subquestion}

%

\noindent\textbf{ID 1}

\begin{tabular}{|p{2.5cm}|p{8.5cm}|p{2.5cm}|}
\hline
\rowcolor{lightgreen}
Source Question & Let $ABC$ be a triangle inscribed in circle $\omega$. Let the tangents to $\omega$ at $B$ and $C$ intersect at point $D$, and let $\overline{AD}$ intersect $\omega$ at $P$. If $AB=5$, $BC=9$, and $AC=10$, $AP$ can be written as the form $\frac{m}{n}$, where $m$ and $n$ are relatively prime integers. Find $m + n$. & 110 \\
\hline
Subquestion 1-1 & In triangle ABC with side lengths AB=5, BC=9, and AC=10, inscribed in a circle $\omega$, suppose we place B at (0,0) and C at (9,0) in the coordinate plane. If A is in the upper half-plane, find the coordinates of A.  & $A(\tfrac{1}{3}, \tfrac{4\sqrt{14}}{3}) $. \\
\hline
Subquestion 1-2 & Points $A(\tfrac{1}{3}, \tfrac{4\sqrt{14}}{3})$, B(0,0), C(9,0) all lie on a circle $\omega$. Find the center and radius of $\omega$. & Center = $(\frac{9}{2},\frac{33\sqrt{14}}{56})$. Radius = $\frac{75\sqrt{14}}{56}$. \\
\hline
Subquestion 1-3 & In the coordinate plane, let $\omega$ be the circle with center $(\tfrac{9}{2}, \tfrac{33\sqrt{14}}{56})$ and radius $\tfrac{75\sqrt{14}}{56}$, and let B(0,0) and C(9,0) be points on $\omega$. Derive the equations of the tangents to $\omega$ at B and at C, and let D be their intersection. Find the coordinates of D. & D$(\frac{9}{2}, \frac{\sqrt{14}}{11})$. \\
\hline
Subquestion 1-4 & In the coordinate plane, let $\omega$ be the circle with center $(\frac{9}{2}, \frac{33\sqrt{14}}{56})$ and radius $\frac{75\sqrt{14}}{56}$, and let B(0,0) and C(9,0) be points on $\omega$. Derive the equations of the tangents to $\omega$ at B and at C, and let D be their intersection. Find the coordinates of D. & AP = $\frac{100}{11}$. \\
\hline
\end{tabular}





\noindent\textbf{ID 5}

\begin{tabular}{|p{2.5cm}|p{8.5cm}|p{2.5cm}|}
\hline
\rowcolor{lightgreen}
Source Question & Let $ABCD$ be a tetrahedron such that $AB=CD= \sqrt{40}$, $AC=BD= \sqrt{80}$, and $BC=AD= \sqrt{89}$. There exists a point $I$ inside the tetrahedron such that the distances from $I$ to each of the faces of the tetrahedron are all equal. This distance can be written in the form $\frac{m \sqrt{n}}{p}$, where $m$, $n$, and $p$ are positive integers, $m$ and $p$ are relatively prime, and $n$ is not divisible by the square of any prime. Find $m+n+p$. & 104 \\
\hline
Subquestion 5-1 & Consider a tetrahedron ABCD with edges AB = CD = $\sqrt{41}$, AC = BD = 60, and BC = AD = $\sqrt{89}$. Using the Cayley-Menger determinant or another appropriate method, find the volume V of this tetrahedron. Give the final value of V. & V = $\frac{160}{3}$ \\
\hline
Subquestion 5-2 & Consider a triangle with side lengths $\sqrt{41}$, $\sqrt{89}$, and 60. Find its area using Heron's formula. Then, for a tetrahedron with four such triangular faces, compute the total surface area A. & 24$\sqrt{21}$. \\
\hline
Subquestion 5-3 & Consider a tetrahedron with volume V = $\frac{160}{3}$ and total surface area A = 24$\sqrt{21}$. Inside this tetrahedron, there is a point I whose distances to the four faces are all equal. Let the common distance (the inradius) be r. Find r in simplest radical form. Suppose r = $\frac{m\sqrt{n}}{p}$, where m, p are relatively prime positive integers, and n is a positive integer not divisible by the square of any prime. & r = $\frac{20\sqrt{21}}{63}$. \\
\hline
\end{tabular}

\vspace{1cm}

\end{document}